%% file: main.tex
\documentclass[twoside,11pt]{article}
\usepackage[preprint]{jmlr2e}
\usepackage{times}
\usepackage{amsmath,mathtools,booktabs,array,microtype,enumitem,needspace,lastpage}
\hypersetup{hidelinks,pdftitle={A Hierarchy of Risk Certificates for Representation-Based Prediction},pdfauthor={Mohit Kumar and Somayeh Kargaran}}
\jmlrheading{}{2026}{1--\pageref{LastPage}}{}{}{}{Mohit Kumar and Somayeh Kargaran}
\ShortHeadings{A Hierarchy of Risk Certificates}{Kumar and Kargaran}
\firstpageno{1}
\allowdisplaybreaks
\newcommand{\clip}{\operatorname{clip}}

\begin{document}
\title{A Hierarchy of Risk Certificates for Representation-Based Prediction}
\author{\name Mohit Kumar \email mohit.kumar@uni-rostock.de\\
\addr University of Rostock, Rostock, Germany\\
Software Competence Center Hagenberg, Hagenberg, Austria
\AND
\name Somayeh Kargaran \email somayeh.kargaran@scch.at\\
\addr Software Competence Center Hagenberg, Hagenberg, Austria}
\maketitle

\begin{abstract}
Representation-based prediction underlies many
machine-learning systems. Yet a fitted predictor's
failure to meet an accuracy requirement does not
establish that its representation-decoder class
cannot meet it. We study optimal population mean
squared error (MSE) over Lipschitz decoders acting
on a fixed representation. For blocks of $b\geq2$
independent and identically distributed (iid)
observations, we minimize total squared error (normalized
by $b-1$) under pairwise Lipschitz constraints. We prove that expected block scores form nondecreasing lower bounds on optimal class
risk and converge to that risk on compact metric
feature spaces, although no fixed block size
provides a uniform relative guarantee. Verified dual lower bounds and iid calibration
observations independent of model development,
with bounded targets and predictions, yield
simultaneous lower confidence bounds on optimal
class risk and upper confidence bounds on
fitted-predictor risk. These bounds can establish
that every admissible decoder exceeds a prescribed
MSE threshold or that an admissible fitted
predictor meets it. Controlled experiments illustrate population-risk
certification; fixed PTB-XL and Bike Sharing
experiments provide deterministic empirical-risk
conclusions. The results support principled
assessment and selection of representation-decoder
configurations. Efficient certification at scale
and extensions to dependent observations remain
open problems.
\end{abstract}


\input{sections/introduction}
\input{sections/related_work}
\input{sections/theory}
\input{sections/representations}
\input{sections/evidence}

\input{sections/discussion}

\section{Data and Code Availability}
The \href{https://github.com/MohitKumarRostock/certified-kahm-representations-forecasting/tree/7710b3eb18ee40859026866169379896a009d556}{\underline{demonstrator repository}}
provides the implementation, scientific documentation,
calibration plan, stored feature and prediction tables,
and complete verification bundle for
Section~\ref{sec:demonstrator}.
Its \texttt{README.md} gives installation commands and
commands for verifying the stored results.
The verifier checks the stored observations, optimization
bounds, confidence endpoints, fitted-table admissibility,
and reported decisions. \texttt{docs/SCIENTIFIC\_REFERENCE.md} gives self-contained
definitions of the risk quantities and assumptions. The \href{https://github.com/MohitKumarRostock/Kernel_Affine_Hull_Koopman_Machine/tree/8baad4931f8a934c4b09612e7e8f90149873a1ca}{\underline{experiment repository}}
provides code, stored inputs, and fitted models for the
other experiments.
\texttt{REPRODUCIBILITY.md} specifies dependencies, commands,
and numerical scope; \texttt{DATA.md} identifies data sources
and attribution. The \texttt{reproduce.py} driver checks \texttt{MANIFEST.json}
and recomputes results from retained records, without
retraining models or collecting new timing measurements.
The deployment check regenerates forecasts from saved models.

\acks{The research reported in this paper has been supported by the GenAISpace project, co-funded by the European Union and Austrian Research Promotion Agency (FFG) under the Digital Europe Programme (Grant Agreement No. 101330264); and the Austrian Ministry for Transport, Innovation and Technology, the Federal Ministry for Digital and Economic Affairs, and the State of Upper Austria through the SCCH competence center INTEGRATE (FFG grant no. 892418), part of the FFG COMET Competence Centers for Excellent Technologies Programme.}

\appendix
\input{sections/resolution}
\input{sections/protocols}

\bibliography{references}
\end{document}

%% file: sections/introduction.tex
\section{Introduction}
\label{sec:introduction}
Many prediction systems transform their inputs
into feature representations and fit decoders
to those features. A fitted predictor may have population mean squared
error (MSE) above a prescribed threshold even when
another decoder in the same representation-decoder
class meets that threshold. We study the infimum of population MSE over
decoders using a fixed representation and
satisfying a prescribed Lipschitz constraint.
We construct a hierarchy of lower bounds on
this optimal class risk. If a certified lower
bound exceeds the prescribed threshold, no
decoder in the class can meet the requirement.
In industrial predictive maintenance, such a
certificate could motivate revising the sensor-history
representation or decoder constraint used to forecast
a machine-condition indicator rather than fitting
further decoders within the excluded class. To illustrate the exclusion mechanism, consider the representation
$\Gamma(x)=(|x_1|,|x_2|)$ of a two-observation history
$x=(x_1,x_2)$.
Suppose the histories $(-1,1)$ and $(1,-1)$ each occur
with probability $1/2$, with respective deterministic
targets $1$ and $-1$.
Both histories have representation $(1,1)$, so every
decoder assigns them the same prediction $c$.
Its population MSE is
\[
\frac{(1-c)^2+(-1-c)^2}{2}=1+c^2\geq1.
\]
No decoder using this representation can meet an MSE
threshold below one.
Even without identical feature values, a Lipschitz constraint
limits prediction differences between nearby feature vectors.
If their targets differ by more than this limit, their two
squared errors cannot both be zero.
Our hierarchy applies these constraints jointly to blocks
of independent and identically distributed (iid) observations.

\emph{Class risk and threshold decisions.}
Let $X$ be the observed history, $Y$ the scalar prediction
target, and $U=\Gamma(X)$ the representation in a feature
space $\mathcal U$ with a prescribed pseudometric $d$.
Assume that $Y$ lies almost surely in a known bounded
interval $I=[y_-,y_+]$. For $\kappa\geq0$, let
$\mathcal G_\kappa(d)$ be the class of measurable decoders
$g:\mathcal U\to I$ satisfying
$|g(u)-g(v)|\leq\kappa d(u,v)$ for all $u,v\in\mathcal U$.
Under the specified evaluation law, define
\[
R(g)=\mathbb E[(Y-g(\Gamma(X)))^2],
\qquad
R_\kappa^*=\inf_{g\in\mathcal G_\kappa(d)}R(g).
\]
The optimal class risk $R_\kappa^*$ depends on the
representation, feature pseudometric, Lipschitz bound,
and evaluation law. It therefore measures the capability
of the specified representation-decoder class, not the
predictive information in the representation alone. Let $\widehat g:\mathcal U\to I$ be a measurable
fitted decoder constructed using development data
and fixed before calibration. For prespecified
candidate classes and fitted decoders, we use
independent iid calibration observations to
construct simultaneous lower confidence bounds
on optimal class risk and upper confidence
bounds on fitted-predictor risk. These bounds
incorporate conservative numerical estimates
and prespecified failure-probability allocations. Writing the bounds as $L$ and $U_{\rm fit}$, Theorem~\ref{thm:certificate} establishes
\[
L\leq R_\kappa^*,
\qquad
R(\widehat g)\leq U_{\rm fit}
\]
simultaneously with probability at least
$1-\delta$, without requiring $\widehat g$
to satisfy the prescribed Lipschitz constraint.
If $\widehat g\in\mathcal G_\kappa(d)$, these
inequalities additionally give
\[
L\leq R_\kappa^*
\leq R(\widehat g)\leq U_{\rm fit}.
\]
For a prespecified threshold $\tau$, the condition
$L>\tau$ excludes every decoder in the class.
If $\widehat g$ is admissible, $U_{\rm fit}\leq\tau$
establishes that the class meets the requirement.
When neither conclusion is established, the
certification decision remains unresolved. For a fixed dataset, verified optimization bounds concern the minimum empirical MSE on that dataset (Proposition~\ref{prop:fixed-data-risk}).

\emph{The block hierarchy.}
For a block of $b\geq2$ iid feature--target observations,
let $T_b$ be the minimum total squared error under all
pairwise Lipschitz constraints, divided by $b-1$.
Writing $\mathcal W_b=\mathbb E[T_b]$, we prove
\[
0\leq\mathcal W_2\leq\mathcal W_3\leq\cdots\leq R_\kappa^*.
\]
On compact metric feature spaces,
$\mathcal W_b\to R_\kappa^*$ as $b\to\infty$
(Theorem~\ref{thm:hierarchy}). Thus, for each fixed
problem and each $\tau<R_\kappa^*$, some finite block
size has $\mathcal W_b>\tau$. This convergence does not
provide a uniform guarantee at any fixed block size:
$\mathcal W_b/R_\kappa^*$ can be arbitrarily small
across problems with $R_\kappa^*>0$
(Proposition~\ref{prop:limits}).

\emph{Computation and calibration.}
Evaluating the dual objective using numerical
multipliers associated with selected pairwise
Lipschitz constraints provides lower bounds
on block minima and minimum empirical risk. These bounds remain valid without exact
optimization (Proposition~\ref{prop:dual}). We use these bounds in two
calibration constructions. The first calibrates the
expected block score from independent disjoint blocks
using~\eqref{eq:Lblock}; the second calibrates optimal class
risk from a lower bound on minimum empirical risk over
all $n$ calibration observations using~\eqref{eq:Lemp}.
The second construction can reuse block computations
through \eqref{eq:reuse}, while retaining sample size
$n$ in its confidence bound. In contrast, the first
uses $\lfloor n/b\rfloor$ independent blocks, so its
confidence bound need not improve as $b$ increases. Before examining the calibration observations, we fix
the candidate classes, fitted predictors, types of
confidence bounds to be reported, and their
failure-probability allocations. For direct block
calibration, we also fix the block sizes and partitions. The simultaneous guarantees
support threshold decisions and bounds on the minimum
assigned cost among prespecified candidate classes
(Corollary~\ref{cor:decisions}).

\emph{Relation to prior work.}
Existing work addresses Lipschitz regression
\citep{gottlieb2017regression}, lower bounds based
on sample-average optima \citep{mak1999montecarlo},
and confidence bounds on optimal population risk
\citep{guigues2017confidence,muller2025falsification}.
Let $a_b$ denote the minimum empirical MSE over
a block of $b$ observations. Although
$\mathbb E[a_b]$ is known to be nondecreasing
with $b$ under iid sampling, this result does
not establish monotonicity of our normalized
block score $T_b=ba_b/(b-1)$, because the factor
$b/(b-1)$ decreases with $b$.
We prove that the expected normalized block scores
are nevertheless nondecreasing and converge to
optimal population class risk on compact metric
feature spaces. We also establish that no fixed
block size provides a uniform relative lower bound
and develop class-risk certificates using verified
optimization bounds and statistical calibration.

\emph{Empirical evaluation and scope.}
We evaluate representation--decoder classes based on
Kernel Affine Hull Machines (KAHMs), raw histories,
and Nystr\"om features. A finite-state experiment
with a known evaluation distribution compares
calibrated risk bounds with optimal population
class risks. It also shows that a fitted predictor
can meet the MSE threshold while violating the
prescribed Lipschitz constraint. A separate
feature-augmentation experiment examines how
representation choice and calibration sample size
affect class-exclusion and predictor-feasibility
decisions. For the fixed PTB-XL dataset, a computed lower bound
on minimum empirical MSE exceeds the prescribed
threshold for the selected two-coordinate KAHM class
with $\kappa=1$. Consequently, no decoder in this
class can meet the empirical MSE requirement, provided
that the numerical upper bounds on feature distances are valid. For the fixed Bike Sharing dataset, feasible prediction
vectors provide upper bounds on minimum empirical MSE
that are below the corresponding thresholds for all
54 decoder classes formed from three prediction targets and 18 feature representations. Consequently, every
class contains an admissible decoder that meets its
empirical MSE requirement. However, in 39 of these
classes, the empirical MSE of each of the two fitted
predictors (the clipped affine ridge predictor and
the clipped Lipschitz-extension predictor) exceeds
the corresponding threshold. This finding shows that evaluating only fitted predictors can lead to the incorrect rejection of decoder classes
that contain admissible predictors meeting the empirical
MSE requirement. The PTB-XL and Bike Sharing
results concern empirical MSE on the evaluated
datasets and do not establish population-risk
guarantees. 

%% file: sections/related_work.tex
\section{Related Work}
\label{sec:related}

\emph{Representations and prediction-risk bounds.}
Kernel Affine Hull Machines (KAHMs) construct features from
local reconstruction models
\citep{kumar2024kahm,kumar2025collaborative,
kumar2026kernelaffinehullmachines}.
\citet{kumar2026limits} derive lower confidence bounds on
the minimum population root mean squared error when
future feature vectors are predicted by linear maps
with a prescribed spectral-norm bound. Their bounds
use variation in future features among states sharing
a reference label. In contrast, we study prediction of a common bounded
scalar target from different fixed representations.
We consider Lipschitz decoder classes defined by
feature pseudometrics and derive a hierarchy of
lower bounds on their optimal population MSE.
The hierarchy is based on blockwise optimization
under joint Lipschitz constraints, rather than
bounds derived from variation in future feature
vectors.

\emph{Lipschitz regression and pairwise error bounds.}
\citet{gottlieb2017regression} develop methods for
Lipschitz regression in metric spaces, including
approximate Lipschitz extensions and finite-sample
prediction guarantees. In representation learning,
\citet[Lemma~2]{akash2021inverse} use a Lipschitz
constraint to lower-bound the MSE of a predictor of
an auxiliary continuous variable. Their bound relates differences in target values to the
prediction differences permitted by the Lipschitz
constraint. Our block optimization extends this
pairwise comparison by enforcing the Lipschitz
constraints jointly across all observations in a block.
We establish the resulting hierarchy of expected
lower bounds on optimal class risk
(Theorem~\ref{thm:hierarchy}).

\emph{Sample-average minima and empirical variance.}
For stochastic optimization under iid sampling,
\citet{mak1999montecarlo} show that the expected
minimum sample-average objective is a lower bound
on the population optimum and is nondecreasing with
sample size. If $a_b$ is the minimum empirical MSE
for a block of $b$ observations, our normalized
block score is $T_b=ba_b/(b-1)$. The factor
$b/(b-1)$ decreases with $b$, so monotonicity of
$\mathbb E[a_b]$ does not by itself establish
monotonicity of $\mathbb E[T_b]$. We prove the
latter property directly and characterize both the
convergence of $\mathbb E[T_b]$ to optimal class
risk and the limitations of any fixed block size
(Theorem~\ref{thm:hierarchy} and
Proposition~\ref{prop:limits}). \citet{belomestny2022variance} analyze the excess
population variance of functions chosen by empirical
variance minimization. Our block score also equals
the minimum sample variance of the residuals
$Y_i-g(U_i)$ over admissible Lipschitz decoders,
using denominator $b-1$ (proof of
Theorem~\ref{thm:hierarchy}). We use the expected
minimum sample variance to lower-bound optimal
population MSE, rather than to bound the excess
variance of an empirically selected decoder.

\emph{Confidence bounds on optimal class risk.}
\citet{guigues2017confidence} develop non-asymptotic
confidence bounds on optimal values in stochastic
optimization. \citet{muller2025falsification}
investigate lower bounds on minimum model-class
risk and establish fundamental limitations of
distribution-free empirical model falsification. We consider prescribed representation-decoder
classes with bounded squared loss and Lipschitz
constraints. Our normalized block optimization
provides a hierarchy of lower bounds on optimal
population class risk. Verified evaluation of
the dual objective gives lower bounds on block
and full-sample empirical minima without
requiring exact optimization. Under independent
iid calibration, Theorem~\ref{thm:certificate}
converts these numerical bounds into simultaneous
lower confidence bounds on optimal population
class risk.

\emph{Risk control and candidate-class selection.}
\citet{angelopoulos2025ltt} use multiple testing to
control the risks of selected prediction procedures,
while \citet{jiang2015abstraction} use statistical
tests to select state abstractions in model-based
reinforcement learning. Our class-exclusion decision
instead asks whether the minimum population MSE over
a prescribed decoder class exceeds the required
threshold. An upper confidence bound on the risk of
an admissible fitted decoder can establish that its
class meets the threshold. For finitely many
prespecified candidate classes with assigned costs,
the simultaneous bounds also give lower and upper
bounds on the minimum cost of meeting the threshold,
without requiring the classes to be nested
(Corollary~\ref{cor:decisions}).

%% file: sections/theory.tex
\section{Prediction Risk and Representation Adequacy}
\label{sec:object}

\emph{Prediction task and evaluation law.}
Let $X\in\mathcal X$ denote the observed history available at prediction time, and let $Y$ denote the scalar future observation to be predicted. A measurable representation $\Gamma:\mathcal X\to\mathcal U$ maps the history to a feature vector $U=\Gamma(X)$, where $\mathcal U$ is the feature space. Model development specifies the representation $\Gamma$, a feature pseudometric $d$, and a Lipschitz bound $\kappa\geq 0$ that constrains the admissible decoders defined below. We call the specified joint distribution of the prediction-time history $X$ and its future target $Y$ the \emph{evaluation law}. A decoder's population risk is its expected squared prediction error under this law. The blockwise population-risk results use independent
and identically distributed feature-target observations
from the specified joint distribution. The confidence
bounds in Section~\ref{sec:calibration} additionally
require the evaluation observations to be independent
of model-development data. The representation, feature
metric, Lipschitz bound, and fitted predictor are
selected using the development data and treated as
fixed when computing the confidence bounds.

\emph{Allowed predictors and their risk.}
We equip the feature space $\mathcal U$ with a measurable, finite-valued pseudometric $d$.
It is nonnegative, symmetric, satisfies the triangle inequality, and obeys $d(u,u)=0$. Unlike a metric, a pseudometric may assign zero distance to distinct feature values. We assume that the target $Y$ lies almost surely in a known finite interval $I=[y_-,y_+]$. 
Let $D=y_+-y_->0$ and $B=D^2$. We exclude $D=0$, for which the target and all admissible predictions are constant and every squared prediction error is zero. For a prescribed Lipschitz bound $\kappa\geq 0$, we define the admissible decoder class and its prediction risks by
\begin{align}
\mathcal G_\kappa(d)
&=\{g:\mathcal U\to I\text{ measurable}:
|g(u)-g(v)|\leq\kappa d(u,v)
\ \text{for all }u,v\},
\label{eq:class}\\
R(g)&=\mathbb E[(Y-g(\Gamma(X)))^2],\quad  R^*_\kappa=\inf_{g\in\mathcal G_\kappa(d)}R(g).
\label{eq:risk}
\end{align}

The Lipschitz constraint requires every admissible decoder to assign the same prediction to feature values at zero distance. Here, $R(g)$ denotes the population mean squared error of an individual decoder, whereas $R^*_\kappa$ is the infimum of the population risks over the entire admissible class. We assume that all infima and statistics used below are measurable. Restricting decoder outputs to $I$ does not increase the minimum attainable risk.
To see this, define $\operatorname{clip}_I(z) =\min\{y_+,\max\{y_-,z\}\}$. Clipping the outputs of any real-valued Lipschitz decoder to $I$ preserves its Lipschitz constraint and cannot increase squared error because $Y\in I$. Because both $Y$ and $g(U)$ lie in $I$, the squared prediction error is bounded by $B=D^2$. This bound is used to construct the confidence bounds in Section~\ref{sec:calibration}. Moreover, the constant decoder
$g(u)=\mathbb E[Y]$ belongs to $\mathcal G_\kappa(d)$ and has population risk $\operatorname{Var}(Y)$.
Consequently,
\[
0\leq R^*_\kappa
\leq \operatorname{Var}(Y)
\leq B/4,
\]
where the final inequality follows from the bounded range of $Y$. The optimal risk $R_\kappa^*$ depends on the representation $\Gamma$, the feature pseudometric $d$, the Lipschitz bound $\kappa$, and the specified evaluation law. It therefore characterizes the predictive capability of a representation together with its admissible decoder class, rather than the representation alone.

\emph{Three sources of unavoidable class error.}
The optimal class risk reflects three sources of prediction error: uncertainty about the future target, information lost through the representation, and restrictions imposed by the decoder class. To distinguish these sources, define $M_X=\mathbb E[Y\mid X]$ and $M_U=\mathbb E[Y\mid U]$, where $U=\Gamma(X)$. Under squared loss, $M_X$ is the optimal prediction based on the full available history, whereas $M_U$ is the optimal prediction based only on its representation, without imposing the Lipschitz constraint. For any admissible decoder $g$, the prediction error can be written as
\[
Y-g(U)=(Y-M_X)+(M_X-M_U)+(M_U-g(U)).
\]
The first residual has conditional mean zero given $X$, and the second has conditional mean zero given $U$. Consequently, all cross terms vanish when taking the expected squared error. Taking the infimum over admissible decoders gives
\begin{equation}
 R^*_\kappa
 =\underbrace{\mathbb E(Y-M_X)^2}
      _{\text{future uncertainty}}
  +\underbrace{\mathbb E(M_X-M_U)^2}
      _{\text{conditional-mean information loss}}
  +\underbrace{\inf_{g\in\mathcal G_\kappa(d)}
       \mathbb E(M_U-g(\Gamma(X)))^2}
      _{\text{Lipschitz approximation error}}.
 \label{eq:decomposition}
\end{equation}
The first term measures the prediction error that remains even when the full history is
available. The second measures the additional error caused by replacing the history with its representation, specifically the loss of information about the conditional mean of $Y$. The third measures the smallest additional error introduced by restricting the decoder to the prescribed Lipschitz class. All three terms are nonnegative. A lower bound on their sum
does not identify which source of error prevents the class from meeting an accuracy requirement.

\section{A Hierarchy of Blockwise Risk Lower Bounds}
\label{sec:hierarchy}
\subsection{A Lower Bound from Pairs of Observations}
Consider two feature-target observations $(U,Y)$ and $(U',Y')$. For any admissible decoder $g$, the Lipschitz constraint requires
\[
|g(U)-g(U')|\leq\kappa d(U,U').
\]
If the difference between the targets exceeds this limit, the decoder cannot predict both
targets exactly. The difference $|Y-Y'|-\kappa d(U,U')$, when positive, measures
the part of the target separation that the predictions cannot reproduce. This discrepancy provides a lower bound on the combined squared prediction error of the two observations. We first consider independent draws from the evaluation law. We also allow paired observations to share a common context, represented by a random variable $A$. Conditional on $A$, the two feature-target pairs $(U,Y)$ and $(U',Y')$ are independent draws from the same conditional distribution. We require both $(U,Y)$ and $(U',Y')$ to have the same marginal distribution as $(\Gamma(X),Y)$ under the specified evaluation law. The decoder uses only the
representation $U$, not the auxiliary variable $A$, and the same decoder applies across all
contexts. Taking $A$ to be constant recovers two independent draws from the evaluation law. For example, we may pair history-target observations taken at the same time index in two independently generated episodes. Conditional on that index, the pairs are independent and identically distributed, provided the episodes follow the same generating process. The sampling procedure must preserve the specified marginal evaluation law. Define $(z)_+=\max\{z,0\}$. The population pair score is
\begin{equation}
 W_{\kappa,A}
 =\frac12\mathbb E\bigl(
 |Y-Y'|-\kappa d(U,U')
 \bigr)_+^2.
 \label{eq:pair}
\end{equation}
The score is positive only when the target difference exceeds the maximum prediction difference permitted by the Lipschitz constraint.

\begin{proposition}[Pairwise risk lower bound]
\label{prop:pair}
Let $A$ be an auxiliary random variable such that, conditional on $A$, the feature-target pairs $(U,Y)$ and $(U',Y')$ are independent and identically distributed. Suppose that each pair has the same marginal distribution as $(\Gamma(X),Y)$ under the specified evaluation law. Then
\[
W_{\kappa,A}\leq R_\kappa^*.
\]
\end{proposition}
\begin{proof}
Fix an admissible decoder $g$ and define the prediction residuals $r=Y-g(U)$ and $r'=Y'-g(U')$. Since $Y-Y'=(r-r')+(g(U)-g(U'))$, the triangle inequality and the Lipschitz
constraint give
\[
\begin{aligned}
|Y-Y'|
&\leq |r-r'|+|g(U)-g(U')|
&\leq |r-r'|+\kappa d(U,U').
\end{aligned}
\]
Consequently,
\[
\bigl(|Y-Y'|-\kappa d(U,U')\bigr)_+
\leq |r-r'|.
\]
Squaring and taking expectations gives
\[
W_{\kappa,A}
\leq\frac12\mathbb E(r-r')^2.
\]
Conditional independence and identical distribution
of $r$ and $r'$ given $A$ imply
\[
\mathbb E[(r-r')^2\mid A]
=2\operatorname{Var}(r\mid A).
\]
Therefore, using the conditional variance identity and $\mathbb E[r^2]=R(g)$, we obtain
\[
\begin{aligned}
W_{\kappa,A}
\leq \frac12\mathbb E(r-r')^2
&=\mathbb E\operatorname{Var}(r\mid A) 
&=R(g)-\mathbb E\bigl[
    (\mathbb E[r\mid A])^2\bigr]
&\leq R(g).
\end{aligned}
\]
Taking the infimum over $g\in\mathcal G_\kappa(d)$ proves the claim.
\end{proof}

\subsection{Block Scores, Monotonicity, and Convergence}
The pairwise bound provides a lower bound on optimal class risk using two observations. We extend this construction to groups of
$b\geq 2$ observations, called \emph{blocks}, by requiring the predictions for all observations in a block to satisfy the Lipschitz constraints simultaneously. Let $(U_i,Y_i)_{i=1}^b$ be independent and identically distributed observations from the
feature-target distribution induced by the specified evaluation law. Let $v=(v_1,\ldots,v_b)\in\mathbb R^b$, where $v_i$ denotes the predicted value assigned to
feature value $U_i$. We define the block score as the minimum total squared prediction error
subject to all pairwise Lipschitz constraints, normalized by $b-1$:
\begin{equation}
T_b=\frac{1}{b-1}
\min_{v\in\mathbb R^b}
\left\{
\sum_{i=1}^b(Y_i-v_i)^2:
|v_i-v_j|\leq\kappa d(U_i,U_j)
\quad(1\leq i<j\leq b)
\right\},
\quad
\mathcal W_b=\mathbb E[T_b].
\label{eq:block}
\end{equation}
The block score $T_b$ is computed from one random block of iid observations, and $\mathcal W_b$ is its expectation. The feasible set in \eqref{eq:block} is nonempty because constant prediction vectors satisfy all pairwise Lipschitz
constraints. It is closed and convex because each constraint is equivalent to two linear inequalities. The squared-error objective therefore attains a unique minimum over this set. Restricting $v$ to $I^b$ does not change the minimum:
clipping each $v_i$ to $I$ preserves the pairwise Lipschitz constraints and cannot increase squared error because $Y_i\in I$.

\emph{Why the normalization uses $b-1$.}
Let $a_b$ denote the minimum average squared error over a block of $b$ observations, so $T_b=ba_b/(b-1)$. To explain the normalization, fix an admissible decoder $g$ and define $r_i=Y_i-g(U_i)$ and $\bar r=b^{-1}\sum_{i=1}^b r_i$. Adding $\bar r$ to every prediction $g(U_i)$ preserves the pairwise Lipschitz constraints and yields residuals $r_i-\bar r$. 
The shifted predictions are feasible in \eqref{eq:block}, and their normalized squared error equals the sample variance of the original residuals. Since $T_b$ is the minimum normalized squared error over all feasible predictions, it cannot exceed this sample variance. For any fixed admissible decoder $g$, the residuals $r_1,\ldots,r_b$ are iid. Their sample variance, computed with denominator $b-1$, is therefore an unbiased estimator of $\operatorname{Var}(Y-g(U))$. Since this bound holds for every admissible decoder $g$, taking the infimum gives $\mathcal W_b\leq R^*_\kappa$. Subtracting $\bar r$ from each residual $r_i$ is used only to establish the bound $\mathcal W_b\leq R^*_\kappa$; it is not part of computing $T_b$ in \eqref{eq:block}.
Although the residual sample variance is unbiased for $\operatorname{Var}(Y-g(U))$, the optimized block score $T_b$ need not be unbiased for $R^*_\kappa$. 
\begin{theorem}[Blockwise risk hierarchy]
\label{thm:hierarchy}
Let $(U_i,Y_i)_{i=1}^b$ be iid observations distributed as $(\Gamma(X),Y)$ under the evaluation law specified in Section~\ref{sec:object}, with $Y_i\in I$
almost surely. For the block scores defined in \eqref{eq:block}, the following hold:
\begin{enumerate}
\item For $b=2$,
\[
T_2=\frac12\bigl(
|Y_1-Y_2|-\kappa d(U_1,U_2)
\bigr)_+^2.
\]
Consequently, $\mathcal W_2=W_{\kappa,A}$ when $A$ is constant.
\item The block score is bounded, and its expectation does not exceed the optimal class risk:
\begin{equation}
C_b=\frac{B\lfloor b^2/4\rfloor}{b(b-1)},
\qquad
0\leq T_b\leq C_b,
\qquad
\mathcal W_b\leq R^*_\kappa.
\label{eq:range}
\end{equation}
\item For a subset $S\subseteq\{1,\ldots,b\}$ with $|S|=m$, let $T_m(S)$ denote the block score defined in \eqref{eq:block}, computed using only the observations indexed by $S$ and the normalization $1/(m-1)$. For $2\leq m\leq b$, the average score over all subsets of size $m$ satisfies
\begin{equation}
\binom{b}{m}^{-1}
\sum_{\substack{S\subseteq\{1,\ldots,b\}\\|S|=m}}
T_m(S)\leq T_b.
\label{eq:subset}
\end{equation}
Consequently, $\mathcal W_2\leq\mathcal W_3\leq\cdots \leq R^*_\kappa$.
\item If $(\mathcal U,d)$ is a compact metric
space, then
$\mathcal W_b\uparrow R^*_\kappa$
as $b\to\infty$.
\end{enumerate}
\end{theorem}

\begin{proof}
\emph{Pair score.} For $b=2$, let $\Delta=Y_1-Y_2$ and $L=\kappa d(U_1,U_2)$. The squared-error objective can be written as
\[
\begin{aligned}
&(Y_1-v_1)^2+(Y_2-v_2)^2
&=\frac12\bigl((Y_1+Y_2)-(v_1+v_2)\bigr)^2+\frac12\bigl(\Delta-(v_1-v_2)\bigr)^2.
\end{aligned}
\]
If $|\Delta|\leq L$, choosing $v_1=Y_1$ and $v_2=Y_2$ satisfies the Lipschitz constraint and gives zero squared error. If $|\Delta|>L$, the prediction difference $v_1-v_2$ closest to $\Delta$ is $L$ when $\Delta>L$ and $-L$ when $\Delta<-L$. With $v_1+v_2=Y_1+Y_2$, the first term vanishes, while the second equals $\frac12(|\Delta|-L)^2$. Combining this with the zero-error case (i.e. 
$|\Delta|\leq L$) gives
\[
T_2=\frac12(|\Delta|-L)_+^2
=\frac12\bigl(
|Y_1-Y_2|-\kappa d(U_1,U_2)
\bigr)_+^2.
\]
Taking expectations over two iid observations gives $\mathcal W_2=W_{\kappa,A}$ when $A$ is constant, proving the first assertion.

\emph{Risk bound and range.}
Fix $g\in\mathcal G_\kappa(d)$, and define $r_i=Y_i-g(U_i)$ and $\bar r=b^{-1}\sum_{i=1}^b r_i$. Consider the shifted predictions $v_i=g(U_i)+\bar r$. Adding the same constant to every prediction preserves the pairwise Lipschitz constraints, so these predictions are feasible in \eqref{eq:block}. The shifted predictions may lie outside $I$, but they remain feasible because \eqref{eq:block} allows prediction vectors in $\mathbb R^b$
subject to the pairwise Lipschitz constraints. For the shifted predictions, the residual at observation $i$ is
$Y_i-v_i=r_i-\bar r$. Therefore,
\[
T_b\leq\frac{1}{b-1}
\sum_{i=1}^b(r_i-\bar r)^2.
\]
For fixed $g$, the residuals are iid because the observations are iid. The expectation of their sample variance therefore equals
$\operatorname{Var}(Y-g(U))$. Taking expectations gives
\[
\mathcal W_b
\leq \operatorname{Var}(Y-g(U))
\leq R(g),
\]
where the second inequality follows from $\operatorname{Var}(Y-g(U))=R(g)-(\mathbb E[Y-g(U)])^2$. Since this bound holds for every $g\in\mathcal G_\kappa(d)$, taking the infimum yields $\mathcal W_b\leq R^*_\kappa$. To establish the range bound, let $\bar Y=b^{-1}\sum_{i=1}^b Y_i$ and choose $v_i=\bar Y$ for every $i$. This constant prediction vector satisfies all pairwise Lipschitz constraints. Therefore,
\[
\begin{aligned}
0\leq T_b
&\leq \frac{1}{b-1}
\sum_{i=1}^b(Y_i-\bar Y)^2
&=\frac{1}{b(b-1)}
\sum_{i<j}(Y_i-Y_j)^2.
\end{aligned}
\]
The sample variance on the right is convex in each $Y_i$, so its maximum over $I^b$ is attained when every target equals an endpoint of $I=[y_-,y_+]$. If $k$ targets equal $y_+$ and the remaining $b-k$ equal $y_-$, each upper-endpoint target forms a pair with each of the $b-k$ lower-endpoint targets. Thus, exactly $k(b-k)$ pairs have squared difference $B=(y_+-y_-)^2$, while all other pairs have zero squared difference.
Consequently,
\[
\begin{aligned}
0\leq T_b
&\leq \max_{k\in\{0,\ldots,b\}}
\frac{B k(b-k)}{b(b-1)}.
\end{aligned}
\]
Since $k$ counts the targets equal to $y_+$, it is an integer satisfying $0\leq k\leq b$. The product $k(b-k)$ is maximized when $k$ is closest to $b/2$, namely when $k=\lfloor b/2\rfloor$ or $k=\lceil b/2\rceil$. Therefore,
\[
\begin{aligned}
0\leq T_b
&\leq \frac{B\lfloor b^2/4\rfloor}{b(b-1)}
=C_b.
\end{aligned}
\]
This establishes the range bound in \eqref{eq:range}.

\emph{Subset comparison.}
Let $v=(v_1,\ldots,v_b)$ minimize the full-block objective in \eqref{eq:block}, and define $r_i=Y_i-v_i$. Adding any constant $c$ to every $v_i$ preserves the pairwise Lipschitz constraints and changes the sum of squared residuals to $\sum_{i=1}^b(r_i-c)^2$. This expression is minimized at $c=b^{-1}\sum_{i=1}^b r_i$. Since $v$ already minimizes the full-block objective, the minimizing shift must be $c=0$. Hence $\sum_{i=1}^b r_i=0$. Using the pairwise identity for squared deviations, we obtain
\[
T_b=\frac{1}{b-1}\sum_{i=1}^b r_i^2
=\frac{1}{b(b-1)}
\sum_{i<j}(r_i-r_j)^2.
\]
Now fix a subset $S\subseteq\{1,\ldots,b\}$ with $|S|=m$, and let $\bar r_S=m^{-1}\sum_{i\in S}r_i$. The predictions $v_i+\bar r_S$ for $i\in S$ satisfy all pairwise Lipschitz constraints within $S$. For these shifted subset predictions, the residual at observation $i\in S$ is $Y_i-(v_i+\bar r_S)=r_i-\bar r_S$. Since $T_m(S)$ is the minimum normalized squared error over all feasible predictions for the subset $S$, it cannot exceed the squared error of these shifted predictions. Thus,
\[
\begin{aligned}
T_m(S)
&\leq\frac{1}{m-1}
\sum_{i\in S}(r_i-\bar r_S)^2
&=\frac{1}{m(m-1)}
\sum_{\substack{i<j\\i,j\in S}}
(r_i-r_j)^2.
\end{aligned}
\]
The equality follows from the identity expressing the sample variance in terms
of pairwise squared differences. For fixed indices $i<j$, a subset of size $m$
containing both indices is formed by choosing the remaining $m-2$ indices from the other $b-2$ indices. Thus, each pair $(i,j)$ belongs to exactly $\binom{b-2}{m-2}$ subsets of size $m$. Summing the preceding inequality over all subsets $S\subseteq\{1,\ldots,b\}$ with $|S|=m$ and dividing by their number $\binom{b}{m}$ gives
\[
\begin{aligned}
&\binom{b}{m}^{-1}
\sum_{\substack{S\subseteq\{1,\ldots,b\}\\|S|=m}}
T_m(S)
&\leq
\frac{1}{m(m-1)\binom{b}{m}}
\sum_{\substack{S\subseteq\{1,\ldots,b\}\\|S|=m}}
\sum_{\substack{i<j\\i,j\in S}}
(r_i-r_j)^2.
\end{aligned}
\]
Interchanging the order of summation groups the squared residual differences by each fixed pair of indices $i<j$:
\[
\begin{aligned}
&\sum_{\substack{S\subseteq\{1,\ldots,b\}\\|S|=m}}
\sum_{\substack{i<j\\i,j\in S}}
(r_i-r_j)^2
&=
\sum_{i<j}(r_i-r_j)^2
\sum_{\substack{S\subseteq\{1,\ldots,b\}\\
|S|=m,\ i,j\in S}}1
&=
\binom{b-2}{m-2}
\sum_{i<j}(r_i-r_j)^2.
\end{aligned}
\]
The inner sum counts the subsets of size $m$ containing both indices $i$ and $j$. We use the binomial identity:
\[
\frac{\binom{b-2}{m-2}}{\binom{b}{m}}
=\frac{m(m-1)}{b(b-1)}
\]
Substituting these identities into the averaged inequality gives
\[
\begin{aligned}
\binom{b}{m}^{-1}
\sum_{\substack{S\subseteq\{1,\ldots,b\}\\|S|=m}}
T_m(S)
&\leq
\frac{\binom{b-2}{m-2}}
{m(m-1)\binom{b}{m}}
\sum_{i<j}(r_i-r_j)^2
&=
\frac{1}{b(b-1)}
\sum_{i<j}(r_i-r_j)^2
&=T_b.
\end{aligned}
\]
This establishes \eqref{eq:subset}. Because the observations are iid, each subset
of size $m$ has the same distribution as a
block of $m$ observations. Consequently,
$\mathbb E[T_m(S)]=\mathcal W_m$ for every
such subset $S$. Taking expectations in
\eqref{eq:subset} therefore gives
$\mathcal W_m\leq\mathcal W_b$ for every
$2\leq m\leq b$.

\emph{Convergence under a fixed evaluation law.}
Consider an infinite iid sequence $(U_i,Y_i)_{i\geq1}$ from the evaluation law, with each block formed
from its first $b$ observations. Let $v=(v_1,\ldots,v_b)$ minimize the objective in \eqref{eq:block}. Clipping each $v_i$ to $I$ preserves the Lipschitz constraints and cannot increase squared error. We may therefore assume that $v_i\in I$ for every $i$. Define the decoder
\[
g_v(u)=\operatorname{clip}_I\left(
\min_{1\leq j\leq b}
\{v_j+\kappa d(u,U_j)\}
\right).
\]
For each fixed $j$, the reverse triangle inequality gives
\[
|d(u,U_j)-d(u',U_j)|\leq d(u,u')
\]
for all $u,u'\in\mathcal U$. Hence, $u\mapsto v_j+\kappa d(u,U_j)$ is $\kappa$-Lipschitz. Taking the pointwise minimum of these functions and clipping to $I$ preserve the Lipschitz bound. 
The measurability of $d$ also implies that $g_v$ is measurable. Hence $g_v\in\mathcal G_\kappa(d)$. For every $i$, the constraints in \eqref{eq:block} imply
\[
v_i\leq v_j+\kappa d(U_i,U_j)
\quad\text{for all }j.
\]
Equality holds when $j=i$, since $d(U_i,U_i)=0$. Hence,
\[
\min_{1\leq j\leq b}
\{v_j+\kappa d(U_i,U_j)\}=v_i.
\]
Since $v_i\in I$, clipping leaves this value unchanged. Therefore, $g_v(U_i)=v_i$ for every $i$. Recall that $a_b=(b-1)T_b/b$ denotes the minimum average squared error over the block in \eqref{eq:block}. Since $g_v(U_i)=v_i$ for every $i$, the empirical mean squared error of $g_v$ equals $a_b$. Conversely, for any $g\in\mathcal G_\kappa(d)$, the prediction vector $(g(U_1),\ldots,g(U_b))$ satisfies the pairwise Lipschitz constraints in \eqref{eq:block}. Thus, every admissible decoder produces a feasible block prediction vector, while the optimal block prediction vector is reproduced by the admissible decoder $g_v$. Consequently,
\[
a_b=\min_{g\in\mathcal G_\kappa(d)}
\frac{1}{b}\sum_{i=1}^b
(Y_i-g(U_i))^2.
\]
Every decoder in $\mathcal G_\kappa(d)$ takes values in the bounded interval $I$ and satisfies the same
Lipschitz bound $\kappa$. Thus, $\mathcal G_\kappa(d)$ is uniformly bounded and equicontinuous. Uniform limits of functions in $\mathcal G_\kappa(d)$ remain $I$-valued and $\kappa$-Lipschitz, so the class is also closed
under uniform convergence. Since $(\mathcal U,d)$ is compact, the Arzel\`a-Ascoli theorem implies that $\mathcal G_\kappa(d)$ is compact in the uniform norm, defined by
\[
\|f-g\|_\infty
=\sup_{u\in\mathcal U}|f(u)-g(u)|.
\]
Compactness implies that, for every $\varepsilon>0$, there exists a finite set $\mathcal F_\varepsilon\subseteq
\mathcal G_\kappa(d)$ such that every $g\in\mathcal G_\kappa(d)$ has an $f\in\mathcal F_\varepsilon$ satisfying
$\|g-f\|_\infty\leq\varepsilon$. To obtain approximation errors that converge to zero, set $\varepsilon=1/j$ for integers $j\geq1$ and denote the corresponding finite sets by $\mathcal F_j$. Thus, for every $g\in\mathcal G_\kappa(d)$, some $f\in\mathcal F_j$ satisfies
\[
\|g-f\|_\infty\leq\frac1j.
\]
Define the empirical risk of a decoder $g$ by
\[
\widehat R_b(g)=\frac1b\sum_{i=1}^b
(Y_i-g(U_i))^2.
\]
For any $f,g\in\mathcal G_\kappa(d)$, $u\in\mathcal U$, and $y\in I$, the width $D=y_+-y_-$ of the target interval gives $|y-f(u)|\leq D$ and $|y-g(u)|\leq D$. Therefore,
\[
\begin{aligned}
|(y-f(u))^2-(y-g(u))^2|
&=|f(u)-g(u)|\,|2y-f(u)-g(u)|
&\leq 2D\|f-g\|_\infty.
\end{aligned}
\]
For $g\in\mathcal G_\kappa(d)$ and its approximation $f\in\mathcal F_j$ satisfying $\|g-f\|_\infty\leq 1/j$, the pointwise squared-loss bound established above implies
\[
\begin{aligned}
|\widehat R_b(g)-\widehat R_b(f)|
\leq \frac{2D}{j},\quad 
|R(g)-R(f)|
&\leq \frac{2D}{j},
\end{aligned}
\]
by averaging over the observations and taking expectations under the evaluation law, respectively. The triangle inequality and the preceding risk bounds give
\[
\begin{aligned}
|\widehat R_b(g)-R(g)|
&\leq
|\widehat R_b(g)-\widehat R_b(f)| +
|\widehat R_b(f)-R(f)| +
|R(f)-R(g)|\\
&\leq
|\widehat R_b(f)-R(f)|+\frac{4D}{j}.
\end{aligned}
\]
Since $f\in\mathcal F_j$, taking the supremum over $g\in\mathcal G_\kappa(d)$ yields
\[
\begin{aligned}
&\sup_{g\in\mathcal G_\kappa(d)}
|\widehat R_b(g)-R(g)|
& \leq
\max_{f\in\mathcal F_j}
|\widehat R_b(f)-R(f)|
+\frac{4D}{j}.
\end{aligned}
\]
For each fixed $j$, the strong law of large numbers implies that the maximum over the
finite set $\mathcal F_j$ converges to zero almost surely as $b\to\infty$. Since $j$ ranges over the positive integers, these convergences hold simultaneously for all $j$ with probability one. Taking $b\to\infty$ and then $j\to\infty$ yields
\[
\sup_{g\in\mathcal G_\kappa(d)}
|\widehat R_b(g)-R(g)|
\longrightarrow 0
\quad\text{almost surely}.
\]
Since $a_b$ is the minimum empirical risk and $R^*_\kappa$ is the infimum population risk over the same decoder class, the difference between these optimal values is bounded by the maximum deviation between empirical and population risks:
\[
|a_b-R^*_\kappa|
\leq
\sup_{g\in\mathcal G_\kappa(d)}
|\widehat R_b(g)-R(g)|
\longrightarrow 0
\quad\text{almost surely}.
\]
Since $T_b=ba_b/(b-1)$ and $b/(b-1)\to1$ as $b\to\infty$, it follows that $T_b\to R^*_\kappa$ almost surely. 
By \eqref{eq:range}, $0\leq T_b\leq C_b\leq B/2$ for every $b\geq2$. Thus, $T_b$ is uniformly bounded, and its almost-sure convergence to $R^*_\kappa$ implies, by the dominated
convergence theorem, that
\begingroup\postdisplaypenalty=10000
\[
\mathcal W_b=\mathbb E[T_b]\longrightarrow
R^*_\kappa
\quad\text{as }b\to\infty.
\]
Since $\mathcal W_b$ is nondecreasing in $b$, we conclude that $\mathcal W_b\uparrow R^*_\kappa$.
\endgroup
\end{proof}

\emph{Interpreting the hierarchy.}
Theorem~\ref{thm:hierarchy} establishes that the
expected block score $\mathcal W_b=\mathbb E[T_b]$
is nondecreasing in $b$. This monotonicity holds
in expectation, not necessarily for scores
computed from individual observed blocks.
For any given block, \eqref{eq:subset} guarantees
that its score is at least the average score
over all subsets of a fixed smaller size.
It does not, however, guarantee that the block
score exceeds the score of every such subset.

\subsection{Limits of a Fixed Block Size}
For a fixed block size $b$, the block score can be zero even when the optimal population risk $R^*_\kappa$ is
positive. Proposition~\ref{prop:limits} illustrates two reasons. In the first example, two low-probability feature values have targets that no admissible decoder can fit simultaneously, but a block is unlikely to contain both feature values. In the second example, the target $Y$ is independent of the feature $U$ and therefore has unavoidable prediction error. However, the feature space contains so many well-separated values that a small block is unlikely to contain repeated feature values. When all feature values in a block are distinct, the Lipschitz constraints allow the observed targets to be fitted exactly. In both examples, the expected block score $\mathcal W_b$ can be arbitrarily small relative to $R^*_\kappa$ for any fixed $b$, even though $\mathcal W_b$ converges to $R^*_\kappa$ as $b$ increases for each fixed evaluation law.
\begin{proposition}[Small expected block scores at a fixed block size]\label{prop:limits}
Consider the following two distributions.
\begin{enumerate}
\item Let $(U,Y)$ take the values $(3,0)$, $(0,-1)$, and $(1,1)$ with respective probabilities $1-2\varepsilon$, $\varepsilon$, and $\varepsilon$, where $0<\varepsilon<1/2$. Take $I=[-1,1]$, $d(u,v)=|u-v|$, and $\kappa=1$. Then
\begin{equation}
 R^*_\kappa=\frac{\varepsilon}{2},
 \qquad
 \mathcal W_b=
 \frac{2b\varepsilon-1+(1-2\varepsilon)^b}{4(b-1)}.
 \label{eq:rare}
\end{equation}
\item Let $U$ be uniformly distributed over $M\geq2$ points in a metric space, with the distance between any two distinct points at least $D/\kappa$, where $D,\kappa>0$. Take $I=[0,D]$, and let $Y$ be independent of $U$, with $\Pr(Y=0)=\Pr(Y=D)=1/2$. Then
\begin{equation}
 R^*_\kappa=\frac{D^2}{4},
 \qquad
 \mathcal W_b=\frac{D^2}{4(b-1)}
 \left[b-M\left\{1-\left(1-\frac1M\right)^b\right\}\right].
 \label{eq:occupancy}
\end{equation}
\end{enumerate}
For every fixed integer $b\geq2$, the ratio $\mathcal W_b/R^*_\kappa$ tends to zero as $\varepsilon\downarrow0$ in the first example or as $M\to\infty$ in the second.
\end{proposition}
\begin{proof}
For the first distribution, let $v_-$ and $v_+$ denote the predictions at $U=0$ and $U=1$.
The Lipschitz constraint requires $|v_+-v_-|\leq1$, although the corresponding
targets differ by two. Consequently,
\[
\min_{|v_+-v_-|\leq1}
\left\{(v_-+1)^2+(v_+-1)^2\right\}
=\frac12,
\]
attained at $v_-=-1/2$ and $v_+=1/2$. These predictions lie in $I=[-1,1]$, and setting the prediction at $U=3$ to zero satisfies the remaining Lipschitz constraints without introducing additional error. The frequent input contributes zero squared error. Since each rare input has probability $\varepsilon$ and their minimum combined squared error is $1/2$, the minimum population risk is $R^*_\kappa=\varepsilon/2$. For a block of $b$ observations, let $N_-$ and $N_+$ count the occurrences of $U=0$ and $U=1$, respectively, and set $N=N_-+N_+$. When $N>0$, the minimum squared residual sum at these inputs is
\[
\min_{|v_+-v_-|\leq1}
\left\{
N_-(v_-+1)^2+N_+(v_+-1)^2
\right\}
=\frac{N_-N_+}{N}.
\]
The minimum is attained at $v_-=-N_-/N$ and $v_+=N_+/N$. Together with a prediction of zero at $U=3$,
these values satisfy all Lipschitz constraints. When $N=0$, the minimum residual sum is zero. Since $N\sim\operatorname{Binomial}(b,2\varepsilon)$ and $N_-\mid N=n\sim\operatorname{Binomial}(n,1/2)$,
for $n\geq1$ we obtain
\[
\begin{aligned}
\mathbb E[(b-1)T_b\mid N=n]
&=\frac{\mathbb E[N_-(n-N_-)\mid N=n]}{n}
&=\frac{n-1}{4}.
\end{aligned}
\]
Thus, using $\mathbb EN=2b\varepsilon$ and $\Pr(N\geq1)=1-(1-2\varepsilon)^b$,
\[
\begin{aligned}
(b-1)\mathcal W_b
&=\frac{\mathbb EN-\Pr(N\geq1)}{4}
&=\frac{2b\varepsilon-1+(1-2\varepsilon)^b}{4},
\end{aligned}
\]
which proves \eqref{eq:rare}.

For the second distribution, let $u_1,\ldots,u_M$ denote the possible values of $U$. Since any two distinct support points satisfy $\kappa d(u_j,u_k)\geq D$, every assignment of predictions in $I=[0,D]$ satisfies the Lipschitz constraints between these points. Thus, predictions at distinct support points can be optimized independently, while all observations with the same input must receive the same prediction. Let $N_j$ be the number of observations in a block with $U_i=u_j$. For $N_j>0$, the squared residual sum at $u_j$ is minimized by predicting the sample mean $\bar Y_j$ of the corresponding targets. Because these targets are independent and have variance $D^2/4$, for $n\geq1$,
\[
\mathbb E\left[
\sum_{i:U_i=u_j}(Y_i-\bar Y_j)^2
\,\middle|\,N_j=n
\right]
=(n-1)\frac{D^2}{4}.
\]
A support point with $N_j=0$ contributes zero squared error. Let $O_b$ denote the number of distinct inputs
observed in the block. Since
$\sum_{j=1}^M N_j=b$, the expected minimum
squared residual sum, conditional on the input
counts, is
\[
\frac{D^2}{4}
\sum_{j:N_j>0}(N_j-1)
=\frac{D^2}{4}(b-O_b).
\]
Each support point has probability
$(1-1/M)^b$ of being absent from the block.
Consequently,
\[
\mathbb E O_b
=M\left\{1-\left(1-\frac1M\right)^b\right\},
\]
and therefore
\[
\mathcal W_b
=\frac{D^2}{4(b-1)}(b-\mathbb E O_b).
\]
Substituting the expression for $\mathbb E O_b$
gives \eqref{eq:occupancy}. To determine the optimal population risk, recall that $Y$ is independent of $U$, with mean $D/2$ and variance $D^2/4$. Hence, for any admissible
decoder $g$,
\[
\mathbb E[(Y-g(U))^2\mid U]
=\frac{D^2}{4}
+\left(g(U)-\frac D2\right)^2.
\]
The constant decoder $g(u)=D/2$ is admissible
and attains the lower bound $D^2/4$. Thus,
$R^*_\kappa=D^2/4$. Finally, fix $b\geq2$. Expanding the expressions in \eqref{eq:rare} and \eqref{eq:occupancy} gives
\[
\frac{\mathcal W_b}{R^*_\kappa}
=
\begin{cases}
b\varepsilon+O(\varepsilon^2),
& \varepsilon\downarrow0
\quad\text{(first distribution)},\\[2mm]
\dfrac{b}{2M}+O(M^{-2}),
& M\to\infty
\quad\text{(second distribution)}.
\end{cases}
\]
Both ratios therefore converge to zero for every fixed block size $b$.
\end{proof}
Proposition~\ref{prop:limits} shows that the expected block score can substantially underestimate the optimal
population risk at a fixed block size. In the first example, $\mathcal W_b/R^*_\kappa$ tends to zero as $\varepsilon\downarrow0$, although $R^*_\kappa=\varepsilon/2$ also tends to zero. The second example shows that the discrepancy can persist even when the optimal population risk remains strictly positive. For every fixed $b$, $\mathcal W_b\to0$ as $M\to\infty$, while $R^*_\kappa=D^2/4$ remains unchanged. The second example also shows that this discrepancy does not require an increasing feature-space diameter. Indeed, we may choose
$d(u_j,u_k)=D/\kappa$ for all distinct support points. For every $M\geq2$, the resulting metric space is
finite and therefore compact, with the same diameter $D/\kappa$. Thus, even when the feature-space diameter
and target interval are fixed, no fixed block size $b$ guarantees that $\mathcal W_b$ is close to
$R^*_\kappa$ uniformly over these distributions. This does not contradict Theorem~\ref{thm:hierarchy}, which establishes convergence as $b$ increases for each fixed evaluation law on a compact metric feature space. 
\section{Empirical Risk Bounds and Population Confidence Bounds}
\label{sec:calibration}
For a fixed dataset, we formulate the minimum
empirical MSE over the Lipschitz decoder class
$\mathcal G_\kappa(d)$ as a constrained
quadratic optimization problem.
Proposition~\ref{prop:dual} establishes that
the dual objective evaluated at arbitrary
real-valued multipliers provides a lower bound
on the minimum total squared error.
This deterministic bound does not require
exact optimization or sampling assumptions. For iid evaluation observations independent of model
development, we construct lower confidence bounds on
the minimum population risk $R_\kappa^*$ using either
scores computed from independent blocks of observations
or lower bounds on the minimum empirical risk. We also
construct upper confidence bounds on the population
risk $R(\widehat g)$ of fitted predictors selected
independently of the evaluation observations. The
resulting bounds hold simultaneously for a finite
collection of candidate predictor classes and fitted
predictors, with probability at least $1-\delta$
(Theorem~\ref{thm:certificate}). 
\subsection{Lower Bounds for a Fixed Data Set}
For a fixed data set $(u_i,y_i)_{i=1}^n$ with $y_i\in I$, define the minimum empirical mean squared error over the admissible decoder class as
\[
a_n=\inf_{g\in\mathcal G_\kappa(d)}
\frac1n\sum_{i=1}^n(y_i-g(u_i))^2.
\]
Unlike the optimal population risk $R^*_\kappa$, this quantity measures the minimum average squared error on the given observations. To compute $a_n$, let $v_i\in I$ denote the prediction assigned to the observed feature value $u_i$. For these predictions to come from an admissible decoder, they must satisfy the Lipschitz constraints $|v_i-v_j|\leq\kappa d(u_i,u_j)$ for all $i<j$. Conversely, suppose $v=(v_1,\ldots,v_n)\in I^n$ satisfies all pairwise Lipschitz constraints. As shown in the proof of Theorem~\ref{thm:hierarchy}, we can define
a decoder on the entire feature space by
\[
g_v(u)=\operatorname{clip}_I\left(
\min_{1\leq j\leq n}
\{v_j+\kappa d(u,u_j)\}
\right).
\]
The triangle inequality and measurability of $d$ ensure that $g_v$ is measurable and $\kappa$-Lipschitz. Moreover, for each observed feature value $u_i$, the pairwise constraints give $v_i\leq v_j+\kappa d(u_i,u_j)$ for every $j$, with equality when $j=i$. Since $v_i\in I$, clipping does not change this value, so $g_v(u_i)=v_i$. Thus, every prediction vector satisfying the pairwise constraints can be realized by an admissible decoder in $\mathcal G_\kappa(d)$. The empirical squared error depends on a decoder $g$ only through its predictions
$g(u_1),\ldots,g(u_n)$ at the observed feature values. As established above, the prediction vectors produced by admissible decoders are exactly those in $I^n$ satisfying the pairwise Lipschitz constraints. Thus, optimizing over
admissible decoders and optimizing over these feasible prediction vectors yield the same minimum empirical mean squared error. Consequently,
\[
n a_n=
\min_{v\in I^n}
\left\{
\sum_{i=1}^n(y_i-v_i)^2:
|v_i-v_j|\leq\kappa d(u_i,u_j)
\quad\text{for all }i<j
\right\}.
\]
The factor $n$ converts the minimum empirical mean squared error into the minimum sum of squared errors. This equivalence holds for every fixed data set and requires no sampling assumptions. A feasible prediction vector provides an upper bound on $a_n$, because its empirical mean squared error cannot be smaller than the minimum over all feasible prediction vectors. To establish a lower bound on $a_n$, we use Lagrangian duality. The dual formulation represents observations as vertices of a graph, with each edge corresponding to a selected pairwise Lipschitz constraint. Each edge is assigned a signed real multiplier.
Proposition~\ref{prop:dual} shows that the resulting dual objective provides a lower bound on the minimum sum of squared errors $n a_n$. This bound is valid for any selection of edges and any finite values of the multipliers. Dividing the dual value by $n$ therefore gives a lower bound on $a_n$. Optimizing the multipliers can improve this bound but is not required for its validity.

\begin{proposition}[Lower bounds from graph constraints and disjoint subsets]
\label{prop:dual}
Let $a_n$ be the minimum empirical mean squared
error defined above, and write
$y=(y_1,\ldots,y_n)^\top$. Let $G=(\{1,\ldots,n\},E)$ be a graph whose
vertices correspond to the observations.
Each edge $(i,j)\in E$ identifies the pairwise
Lipschitz constraint
$|v_i-v_j|\leq\kappa d(u_i,u_j)$
used in the dual calculation.
An arbitrary orientation of the edges defines
the incidence matrix
$M\in\mathbb R^{n\times |E|}$.
For an edge directed from $i$ to $j$, the
corresponding column of $M$ has $+1$ in row $i$,
$-1$ in row $j$, and zeros elsewhere. For any vector of real-valued edge multipliers
$\lambda=(\lambda_e)_{e\in E}$, let $s=M\lambda$.
Then
\begin{equation}
D_G(\lambda)
=s^\top y
-\kappa\sum_{e=(i,j)\in E}
d(u_i,u_j)|\lambda_e|
-\frac14\|s\|_2^2
\leq n a_n.
\label{eq:dual}
\end{equation}
The inequality holds for any selection of edges
and any multiplier vector $\lambda$.
Using only a subset of the pairwise constraints
may weaken the lower bound but does not
invalidate it. A lower bound on $a_n$ can also be obtained
by combining bounds computed on disjoint subsets
of the observations.
Let $S_1,\ldots,S_K$ be disjoint subsets of
$\{1,\ldots,n\}$.
For each subset $S_k$, let
$\underline A_k\geq0$ be a verified lower bound
on the minimum sum of squared errors over
predictions in $I$ satisfying all pairwise
Lipschitz constraints within that subset.
Then
\begin{equation}
d_n=\frac1n\sum_{k=1}^K\underline A_k
\leq a_n.
\label{eq:reuse}
\end{equation}
The subsets need not cover all observations.
Observations outside these subsets contribute
nothing to the numerator, while the denominator
remains the full sample size $n$.
\end{proposition}

\begin{proof}
We first prove \eqref{eq:dual}. Let $v\in I^n$
satisfy all pairwise Lipschitz constraints. For each selected edge $e=(i,j)\in E$, the corresponding component $(M^\top v)_e$ equals $v_i-v_j$ or $v_j-v_i$, depending on the chosen orientation. Hence, $|(M^\top v)_e|=|v_i-v_j|$. Hence, for any
real-valued multiplier $\lambda_e$,
\[
\lambda_e(M^\top v)_e
\leq |\lambda_e|\,|v_i-v_j|
\leq \kappa d(u_i,u_j)|\lambda_e|.
\]
Summing over the selected edges and using
$s=M\lambda$ gives
\[
s^\top v
\leq \kappa\sum_{e=(i,j)\in E}
d(u_i,u_j)|\lambda_e|.
\]
Thus,
\[
\begin{aligned}
\|y-v\|_2^2
&\geq \|y-v\|_2^2+s^\top v
-\kappa\sum_{e=(i,j)\in E}
d(u_i,u_j)|\lambda_e|
&=\left\|v-y+\frac{s}{2}\right\|_2^2
+D_G(\lambda)
&\geq D_G(\lambda).
\end{aligned}
\]
Thus, $D_G(\lambda)$ is a lower bound on the squared residual sum of every feasible
prediction vector. Taking the minimum
over all feasible vectors gives
$D_G(\lambda)\leq n a_n$.
The argument uses only the constraints
associated with the selected edges, so
the bound remains valid when the graph
does not include every observation pair. 

We next prove \eqref{eq:reuse}. Let $v$ be any prediction vector satisfying all pairwise Lipschitz constraints on the
full data set. Its restriction to each subset
$S_k$ satisfies all pairwise constraints
within that subset. By the definition of
$\underline A_k$,
\[
\sum_{i\in S_k}(y_i-v_i)^2
\geq \underline A_k.
\]
Since the subsets are disjoint, their
squared residual sums contain no repeated
terms. The squared residuals of observations
outside these subsets are nonnegative.
Consequently,
\[
\begin{aligned}
\sum_{i=1}^n(y_i-v_i)^2
&=\sum_{k=1}^K
\sum_{i\in S_k}(y_i-v_i)^2 +
\sum_{i\notin\bigcup_{k=1}^K S_k}
(y_i-v_i)^2
&\geq\sum_{k=1}^K\underline A_k.
\end{aligned}
\]
This inequality holds for every feasible
prediction vector on the full data set.
Minimizing the left-hand side over all
such vectors and dividing by $n$ gives
$d_n\leq a_n$, proving \eqref{eq:reuse}.
\end{proof}
A numerically verified lower bound
$\underline D_G\leq D_G(\lambda)$ on the dual
objective gives
\[
0\leq \underline A:=\max\{0,\underline D_G\}
\leq n a_n.
\]
Conversely, any vector $v\in I^n$ satisfying
all pairwise Lipschitz constraints provides
an upper bound on the minimum empirical mean
squared error:
\[
a_n\leq\frac1n\sum_{i=1}^n(y_i-v_i)^2.
\]
The dual lower bound can be computed using
selected pairwise constraints, whereas a
feasible upper bound must satisfy all pairwise
constraints.

\emph{Combining lower bounds from disjoint blocks.}
Let $S_1,\ldots,S_K$ be disjoint subsets of
a data set containing $n$ observations, with
$|S_k|=b\geq2$. For each block $S_k$, let
$\underline A_k\geq0$ be a verified lower bound
on its minimum total squared error, and define
$\underline T_k=\underline A_k/(b-1)$.
Because the blocks are disjoint, $\sum_{k=1}^K\underline A_k$ is a lower bound on the minimum total squared error over the full data set. By
\eqref{eq:reuse},
\begin{equation}
d_n=\frac1n\sum_{k=1}^K\underline A_k
=\frac{K(b-1)}{n}
\left(\frac1K\sum_{k=1}^K\underline T_k\right)
\leq a_n.
\label{eq:normalization}
\end{equation}
The denominator is the total data set size $n$,
even if the blocks do not cover all observations. 
\subsection{Population Confidence Bounds and Simultaneous Guarantees}
We next construct confidence bounds for the population
mean of iid observations taking values in a known
bounded interval. These bounds use the binary
Kullback-Leibler (KL) divergence, defined for
$q,p\in[0,1]$ by
\[
\operatorname{kl}(q\Vert p)
=q\log\frac{q}{p}
+(1-q)\log\frac{1-q}{1-p}.
\]
Terms with zero numerator are defined to be zero,
whereas a positive numerator divided by zero gives
infinity. For $q\in[0,1]$ and $c>0$, define
\[
\begin{aligned}
k_-(q,c)
&=\min\bigl\{p\in[0,q]:
\operatorname{kl}(q\Vert p)\leq c\bigr\},\\
k_+(q,c)
&=\max\bigl\{p\in[q,1]:
\operatorname{kl}(q\Vert p)\leq c\bigr\}.
\end{aligned}
\]
These functions give the lower and upper endpoints
of the interval of values $p\in[0,1]$ satisfying
$\operatorname{kl}(q\Vert p)\leq c$. Both sets are nonempty because
$\operatorname{kl}(q\Vert q)=0\leq c$.
They are also closed subsets of $[0,1]$
and hence compact, so their minimum
and maximum are attained. 
For fixed $c>0$, both $k_-(q,c)$ and
$k_+(q,c)$ are nondecreasing in $q$.
To see this, observe that, for $p,q\in(0,1)$,
\[
\frac{\partial}{\partial q}
\operatorname{kl}(q\Vert p)
=\log\frac{q(1-p)}{p(1-q)},
\qquad
\frac{\partial}{\partial p}
\operatorname{kl}(q\Vert p)
=\frac{p-q}{p(1-p)}.
\]
These derivatives have opposite signs whenever
$p\ne q$. 
For fixed $c>0$, let $p=p(q)$ denote either
endpoint satisfying
$\operatorname{kl}(q\Vert p)=c$.
Differentiating this equality with respect
to $q$ gives
\[
\frac{\partial\operatorname{kl}}{\partial q}
+
\frac{\partial\operatorname{kl}}{\partial p}
\frac{dp}{dq}=0,
\]
and therefore
\[
\frac{dp}{dq}
=
-\frac{\partial\operatorname{kl}/\partial q}
{\partial\operatorname{kl}/\partial p}>0,
\]
because the two partial derivatives have
opposite signs whenever $p\ne q$.
Thus, both endpoints increase with $q$
for $q\in(0,1)$. Monotonicity at
$q=0$ and $q=1$ follows by continuity. For fixed $q$, increasing $c$ enlarges the set
of values $p$ satisfying
$\operatorname{kl}(q\Vert p)\leq c$.
Consequently, $k_-(q,c)$ is nonincreasing
in $c$, while $k_+(q,c)$ is nondecreasing
in $c$. Let $V_1,\ldots,V_N$ be iid random variables
taking values in $[0,C]$, where $C>0$ and $N\geq1$.
Write
\[
\bar V=\frac1N\sum_{i=1}^N V_i,
\qquad
\mu=\mathbb E[V_1].
\]
For any $0<\alpha<1$, Hoeffding's
bounded-variable inequality
\citep[Theorem~1, Eq.~(2.1)]{hoeffding1963},
applied to $V_i/C$ and $1-V_i/C$,
gives the following one-sided confidence
bounds:
\begin{equation}
\begin{aligned}
\Pr\left\{
C k_-\left(
\frac{\bar V}{C},
\frac{\log(1/\alpha)}{N}
\right)\leq\mu
\right\}
&\geq1-\alpha,\\
\Pr\left\{
\mu\leq
C k_+\left(
\frac{\bar V}{C},
\frac{\log(1/\alpha)}{N}
\right)
\right\}
&\geq1-\alpha.
\end{aligned}
\label{eq:kl}
\end{equation}
Each inequality provides a one-sided confidence
bound on the population mean $\mu$. The
probabilities refer to the randomness of the
iid sample, while $\mu$ is fixed. Each inequality holds separately with probability
at least $1-\alpha$. To ensure that both inequalities
hold simultaneously with probability at least
$1-\delta$, it suffices to use $\alpha=\delta/2$
in each bound. This follows from the union bound,
which limits the probability that either
inequality fails to at most $\delta$
\citep[Theorem~1, equation~(2.1)]{hoeffding1963}. If $C=0$, the random variables are identically zero, so both bounds are zero and no division by $C$ is needed.

\emph{Calibrating the empirical minimum.}
The concentration inequality \eqref{eq:kl}
applies to averages of iid bounded random
variables. For a fixed decoder $g$ independent
of the evaluation data, its empirical risk
is such an average because the individual
prediction losses are iid. In contrast, the empirical minimum $a_n$
is obtained by minimizing over decoders
using the same evaluation observations.
A verified lower bound $d_n\leq a_n$
may also depend jointly on these observations
through the optimization procedure.
Therefore, $d_n$ cannot generally be treated
as an average of iid losses to which
\eqref{eq:kl} applies directly. Nevertheless, $d_n$ is no greater than
the empirical risk of any fixed admissible
decoder. Proposition~\ref{prop:empirical}
uses this inequality, together with
\eqref{eq:kl}, to obtain a lower confidence
bound on the optimal population risk.

\begin{proposition}[Lower confidence bound from an empirical-risk lower bound]
\label{prop:empirical}
Let $(X_i,Y_i)_{i=1}^n$ be $n\geq1$ iid evaluation observations, and let the nonempty predictor class be fixed independently of these observations. For each admissible decoder $g$, let $\ell_g(X,Y)=(Y-g(\Gamma(X)))^2$ denote its squared prediction loss. As established in Section~\ref{sec:object}, this loss is measurable and takes values in $[0,B]$, where
$B=(y_+-y_-)^2>0$ is known. Let $a_n=\inf_g\frac1n\sum_{i=1}^n
\ell_g(X_i,Y_i)$ denote the minimum empirical risk, assumed measurable. Suppose a numerical procedure returns a measurable value $d_n$ satisfying $0\leq d_n\leq a_n$ for every evaluation data set. Then, for any
$0<\alpha<1$,
\begin{equation}
\Pr\left\{
B k_-\left(
\frac{d_n}{B},
\frac{\log(1/\alpha)}{n}
\right)
\leq
\inf_g\mathbb E[\ell_g(X,Y)]
\right\}
\geq1-\alpha.
\label{eq:empirical}
\end{equation}
\end{proposition}

\begin{proof}
Let
\[
r^*=\inf_g\mathbb E[\ell_g(X,Y)]
\]
denote the minimum population risk.
By the definition of the infimum, there exists
a sequence of predictors $g_1,g_2,\ldots$,
chosen independently of the evaluation data,
whose population risks
$r_j=\mathbb E[\ell_{g_j}(X,Y)]$
decrease to $r^*$. For each $g_j$, define its empirical risk as
\[
\widehat r_j
=\frac1n\sum_{i=1}^n
\ell_{g_j}(X_i,Y_i).
\]
Since $d_n\leq a_n$ and $a_n$ is the minimum
empirical risk over the predictor class,
\[
d_n\leq a_n\leq\widehat r_j
\]
holds for every evaluation data set and every
$j$. Let $c=\log(1/\alpha)/n$.
Because $k_-(q,c)$ is nondecreasing in $q$,
\[
B k_-(d_n/B,c)
\leq B k_-(\widehat r_j/B,c).
\]
For each fixed $j$, the losses
$\ell_{g_j}(X_1,Y_1),\ldots,
\ell_{g_j}(X_n,Y_n)$
are iid and bounded by $B$.
The lower confidence bound in \eqref{eq:kl}
therefore gives
\[
\Pr\left\{
B k_-(\widehat r_j/B,c)>r_j
\right\}\leq\alpha.
\]
Since the lower endpoint computed from $d_n$
cannot exceed the one computed from
$\widehat r_j$, it follows that
\[
\Pr\left\{
B k_-(d_n/B,c)>r_j
\right\}\leq\alpha
\]
for every $j$. Finally, define
\[
E_j=\left\{
B k_-(d_n/B,c)>r_j
\right\}.
\]
Since $r_{j+1}\leq r_j$, these events satisfy
$E_j\subseteq E_{j+1}$. Moreover, because
$r_j\downarrow r^*$,
\[
\bigcup_{j=1}^{\infty}E_j
=
\left\{
B k_-(d_n/B,c)>r^*
\right\}.
\]
Indeed, $r_j\geq r^*$ for every $j$, and any
value strictly greater than $r^*$ exceeds
$r_j$ for all sufficiently large $j$.
By continuity of probability for increasing
sequences of events,
\[
\Pr\left\{
B k_-(d_n/B,c)>r^*
\right\}
=
\lim_{j\to\infty}\Pr(E_j)
\leq\alpha.
\]
Thus, with probability at least $1-\alpha$,
\begingroup\postdisplaypenalty=10000
\[
B k_-(d_n/B,c)\leq r^*,
\]
which proves \eqref{eq:empirical}.
\endgroup
\end{proof}
Proposition~\ref{prop:empirical} provides a lower
confidence bound on optimal population class risk
without a term depending on the size or complexity
of the decoder class. Its validity follows from
the inequality $d_n\leq a_n\leq\widehat R_n(g)$
for every admissible decoder $g$, together with
the bounded-loss concentration inequality.
The required uniform loss bound follows from
the prescribed bounds on targets and decoder
predictions.
\begin{theorem}[Simultaneous confidence bounds for optimal and fitted-predictor risk]
\label{thm:certificate}
Let $m=1,\ldots,M$ index prediction problems
specified using development data. For candidate
$m$, let $\Gamma_m$ be its measurable
representation, $\mathcal G_m=
\mathcal G_{\kappa_m}(d_m)$ the class of
measurable $\kappa_m$-Lipschitz decoders under
the feature pseudometric $d_m$ with outputs in
$I=[y_-,y_+]$, and
$\widehat g_m:\mathcal U_m\to I$ a measurable
fitted decoder. We do not require
$\widehat g_m\in\mathcal G_m$.
The representation, feature pseudometric,
Lipschitz bound, target interval, evaluation
distribution, and fitted decoder are fixed
independently of the calibration data. For each candidate, let $(X_i,Y_i)_{i=1}^n$, $n\geq1$, be iid calibration
observations from its evaluation distribution, independent of the
development data, with $Y_i\in I$ almost surely. Define the population
risk, optimal population risk, and minimum empirical risk by
\[
\begin{aligned}
R_m(g)
&=\mathbb E[(Y-g(\Gamma_m(X)))^2],
& R_m^*
&=\inf_{g\in\mathcal G_m}R_m(g),\\
a_{m,n}
&=\inf_{g\in\mathcal G_m}\frac1n
  \sum_{i=1}^n(Y_i-g(\Gamma_m(X_i)))^2.
\end{aligned}
\]
Set $B=(y_+-y_-)^2>0$. Candidate indices are suppressed in the
calibration observations, the interval $I$, and the range bounds $B,C_b$. The confidence bounds to be computed, their failure probabilities,
and all block sizes and block index sets are fixed before examining
the calibration data. For each included block size $2\leq b\leq n$,
the block scores use $K_b=\lfloor n/b\rfloor$ disjoint blocks of $b$
observations. Let $T_{m,b,k}$ be the minimum sum of squared errors
under the pairwise Lipschitz constraints for block $k$, divided by
$b-1$, as in \eqref{eq:block}. Its expectation and range satisfy
\[
\mathcal W_{m,b}=\mathbb E[T_{m,b,k}],
\qquad
0\leq T_{m,b,k}\leq
C_b=\frac{B\lfloor b^2/4\rfloor}{b(b-1)}.
\]
The fitted decoder's squared losses are
$\ell_{m,i}=(Y_i-\widehat g_m(\Gamma_m(X_i)))^2$.
The numerical inputs to each included bound are measurable and
satisfy, for every calibration data set, the corresponding inequalities
\[
0\leq\underline T_{m,b,k}\leq T_{m,b,k},\qquad
0\leq d_{m,n}\leq a_{m,n},\qquad
\ell_{m,i}\leq\overline\ell_{m,i}\leq B.
\]
For the included bounds, with allocated failure probabilities in $(0,1)$, define
\begin{align}
L^{\rm blk}_{m,b}
&=C_b k_-\!\left(
\frac{\frac{1}{K_b}\sum_{k=1}^{K_b}
\underline T_{m,b,k}}{C_b},
\frac{\log(1/\alpha^{\rm blk}_{m,b})}{K_b}
\right),
\label{eq:Lblock}\\
L^{\rm emp}_m
&=B k_-\!\left(
\frac{d_{m,n}}{B},
\frac{\log(1/\alpha^{\rm emp}_m)}{n}
\right),
\label{eq:Lemp}\\
U_m
&=B k_+\!\left(
\frac{\frac1n\sum_{i=1}^n
\overline\ell_{m,i}}{B},
\frac{\log(1/\alpha^{\rm up}_m)}{n}
\right).
\label{eq:upper}
\end{align}
The functions $k_-$ and $k_+$ are the lower and upper endpoints used
in \eqref{eq:kl}. For every candidate $m$, the simultaneous guarantee includes $U_m$. Let $\mathcal B_m$ denote the prespecified set of block sizes for
which $L^{\rm blk}_{m,b}$ is included. With $L^{\rm emp}_m=0$
when that lower confidence bound is omitted, the combined lower
confidence bound is
\[
L_m=\max\!\left(
\{0,L^{\rm emp}_m\}
\cup
\{L^{\rm blk}_{m,b}:b\in\mathcal B_m\}
\right).
\]
If the sum of the allocated failure probabilities
over all candidates and included bounds is at most
$\delta\in(0,1)$, then, conditional on the
development data, with probability at least
$1-\delta$,
\begin{equation}
L_m\leq R_m^*,
\qquad
R_m(\widehat g_m)\leq U_m
\label{eq:separate-confidence}
\end{equation}
hold simultaneously for every candidate $m$. For each candidate whose fitted decoder satisfies
$\widehat g_m\in\mathcal G_m$, the same event
additionally gives
\begin{equation}
L_m\leq R_m^*\leq R_m(\widehat g_m)\leq U_m.
\label{eq:sandwich}
\end{equation}
On the same event, each included block-based lower bound also satisfies
$L^{\rm blk}_{m,b}\leq\mathcal W_{m,b}$.
The confidence bounds may share calibration observations;
independence between their confidence statements is not required.
\end{theorem}

\begin{proof}
Conditional on the development data, the predictor classes and fitted
decoders are fixed. For each included pair $(m,b)$, the exact block
scores $T_{m,b,1},\ldots,T_{m,b,K_b}$ are iid in $[0,C_b]$ with
expectation $\mathcal W_{m,b}$. Since
$\underline T_{m,b,k}\leq T_{m,b,k}$ and $k_-$ is nondecreasing
in its first argument, the lower confidence inequality in
\eqref{eq:kl} gives
\[
\Pr\left\{
L^{\rm blk}_{m,b}>\mathcal W_{m,b}
\,\middle|\,\text{development data}
\right\}
\leq\alpha^{\rm blk}_{m,b}.
\]
Theorem~\ref{thm:hierarchy} establishes
$\mathcal W_{m,b}\leq R_m^*$. For each included bound $L^{\rm emp}_m$, the condition
$0\leq d_{m,n}\leq a_{m,n}$ and Proposition~\ref{prop:empirical}
give
\[
\Pr\left\{
L^{\rm emp}_m>R_m^*
\,\middle|\,\text{development data}
\right\}
\leq\alpha^{\rm emp}_m.
\]
For each candidate $m$, the exact losses
$\ell_{m,1},\ldots,\ell_{m,n}$ of the fitted decoder are iid in
$[0,B]$. Since $\ell_{m,i}\leq\overline\ell_{m,i}$ and $k_+$
is nondecreasing in its first argument, the upper confidence
inequality in \eqref{eq:kl} gives
\[
\Pr\left\{
R_m(\widehat g_m)>U_m
\,\middle|\,\text{development data}
\right\}
\leq\alpha^{\rm up}_m.
\]
The allocated failure probabilities sum to at most $\delta$. The union bound therefore gives conditional probability at least $1-\delta$ that all included lower and upper
confidence bounds hold simultaneously. On this event, the definition of $L_m$ and $R_m^*\geq0$ imply $L_m\leq R_m^*$, while the fitted-risk upper bounds
give $R_m(\widehat g_m)\leq U_m$ for every candidate $m$. This proves \eqref{eq:separate-confidence}. For each candidate with $\widehat g_m\in\mathcal G_m$, the definition of optimal class risk additionally gives $R_m^*\leq R_m(\widehat g_m)$, establishing \eqref{eq:sandwich}. The same event includes $L^{\rm blk}_{m,b}\leq\mathcal W_{m,b}$ for every included pair $(m,b)$. The union bound does not require independence between the confidence statements. Averaging over the development data gives the same unconditional probability guarantee.
\end{proof}
Table~\ref{tab:certification-outcomes} summarizes the
class decisions for a prespecified population MSE threshold $\tau$.
\begin{table}[tbp]
\centering\small
\begin{tabular}{@{}ll@{}}
\toprule
Condition & Class decision\\
\midrule
$L_m>\tau$ & Excluded: no admissible decoder meets the threshold\\
$U_m\leq\tau$, $\widehat g_m\in\mathcal G_m$ & Requirement met by an admissible decoder\\
Otherwise & Unresolved\\
\bottomrule
\end{tabular}
\caption{Class-certification outcomes on the simultaneous
confidence event of Theorem~\ref{thm:certificate}.}
\label{tab:certification-outcomes}
\end{table}
The simultaneous guarantee in Theorem~\ref{thm:certificate}
remains valid for any candidate selected from the prespecified
collection using the calibration results, without an additional
allocation of failure probability. The choice of pairwise Lipschitz constraints used in
\eqref{eq:dual}, their dual multipliers, the optimization steps,
and the disjoint subsets used in \eqref{eq:reuse} may depend
on the calibration data. The confidence bound \eqref{eq:Lemp}
remains valid provided that the resulting $d_{m,n}$ is measurable
and satisfies $0\leq d_{m,n}\leq a_{m,n}$ for every calibration
data set. For any finite collection $d_{m,n}^{(1)},\ldots,d_{m,n}^{(J)}$
satisfying these conditions for the same empirical minimum
$a_{m,n}$,
\[
0\leq\max_{1\leq j\leq J}d_{m,n}^{(j)}\leq a_{m,n}.
\]
Their maximum can therefore replace $d_{m,n}$ in \eqref{eq:Lemp}
with the same allocated failure probability $\alpha^{\rm emp}_m$. For \eqref{eq:Lblock}, the block sizes and block index sets are
fixed before examining the calibration data. Numerical optimization
within these blocks may depend on the calibration observations,
provided that $0\leq\underline T_{m,b,k}\leq T_{m,b,k}$ for every
data set. For fixed $n$, increasing $b$ can reduce the number
$K_b$ of independent block scores. Neither $L^{\rm blk}_{m,b}$
nor $L^{\rm emp}_m$ is guaranteed to exceed the other. The guarantee in Theorem~\ref{thm:certificate} does not cover new
candidates, block sizes, or block index sets introduced after
examining the calibration data. For repeated applications of the
theorem at prespecified sample sizes $n_1,\ldots,n_J$, the included
bounds and their failure probabilities are specified before
calibration. If the total allocated failure probability across
all candidates, bounds, and sample sizes is at most $\delta$,
the union bound gives probability at least $1-\delta$ that all
included bounds hold simultaneously over the $J$ assessments.

\emph{Evaluation units and loss definition.}
Let the calibration data consist of $n$ iid time-series realizations,
independent of the development data. For the lower confidence
bounds, a common prespecified rule is applied separately to each
realization to sample one prediction origin, independently across
realizations. Conditional on the development data, the resulting
history-target pairs are iid. Pairs from multiple origins within
the same realization need not be independent. Let $(X_{i,t},Y_{i,t})$ denote the history-target pair at origin $t$ in realization $i$, and let $p_{i,t}$ be the probability that the same sampling rule selects that origin. The average loss for candidate $m$ on realization $i$ is
\[
V_{m,i}
=\sum_t p_{i,t}
\bigl(Y_{i,t}-\widehat g_m(\Gamma_m(X_{i,t}))\bigr)^2,
\qquad
p_{i,t}\geq0,\quad \sum_t p_{i,t}=1,
\]
where $t$ ranges over the eligible origins in realization $i$.
Conditional on the development data, the $V_{m,i}$ are iid in
$[0,B]$ and satisfy
$\mathbb E[V_{m,i}\mid\text{development data}]
=R_m(\widehat g_m)$. Equation~\eqref{eq:upper} applies with measurable upper bounds
$\overline V_{m,i}$ in place of $\overline\ell_{m,i}$, provided
$V_{m,i}\leq\overline V_{m,i}\leq B$ for every calibration data set.
The sample size $n$ counts independent time-series realizations. The assumption $Y\in I=[y_-,y_+]$ almost surely applies to the
evaluation distribution induced by this sampling rule.
Together with decoder outputs in $I$, it gives the squared-loss
bound $B=(y_+-y_-)^2$. The minimum and maximum targets observed
during development do not establish this assumption for future
observations. If targets outside $I$ are clipped, the bounds concern prediction
error relative to the clipped targets. Clipping decoder outputs
to $I$ leaves the target unchanged. Since capped squared loss is
no larger than the original squared loss, an upper confidence
bound on its expectation does not, by itself, bound ordinary
mean squared error.

\subsection{Numerical Verification of Risk Bounds}
The dual inequality \eqref{eq:dual} holds for
arbitrary real-valued multipliers. We interpret
the stored floating-point multipliers $\lambda$
as exact rational values and set $s=M\lambda$. For each selected edge $e=(i,j)$, let
$d_e^+\geq d(u_i,u_j)$ be a verified upper
bound on the corresponding feature distance.
Since the distance penalty is subtracted in
the dual objective, we obtain
\[
\begin{aligned}
s^\top y
-\kappa\sum_{e\in E}d_e^+|\lambda_e|
-\frac14\|s\|_2^2
&  \leq  D_G(\lambda) &  \leq  n a_n.
\end{aligned}
\]
We evaluate the expression on the left using
exact rational arithmetic when its inputs
are rational. Otherwise, outward-rounded
interval arithmetic provides a verified
lower bound $\underline D_G$ on this
expression. Consequently,
$\underline D_G\leq n a_n$. Because the minimum total squared error is
nonnegative, any downward-rounded value
$d_n$ satisfying
\[
0\leq d_n\leq
\frac{\max\{0,\underline D_G\}}{n}
\leq a_n
\]
is a valid numerical lower bound on minimum
empirical MSE. Sums of verified lower bounds and their normalizations also use
downward rounding. In \eqref{eq:Lblock} and \eqref{eq:Lemp}, numerical evaluation
uses verified lower bounds on
$\frac{1}{K_b C_b}\sum_{k=1}^{K_b}\underline T_{m,b,k}$
and $d_{m,n}/B$, respectively, as the first argument of $k_-$.
In \eqref{eq:upper}, numerical evaluation uses a verified upper
bound on $\frac{1}{nB}\sum_{i=1}^n\overline\ell_{m,i}$
as the first argument of $k_+$. These input bounds lie in $[0,1]$.
Both calculations use an upper bound on
$c=\log(1/\alpha)/N$, since $k_-$ is nonincreasing and $k_+$
is nondecreasing in $c$.
Here $N=K_b$ in \eqref{eq:Lblock} and $N=n$ in
\eqref{eq:Lemp} and \eqref{eq:upper}.
Verified lower bounds on $k_-$ and upper bounds on $k_+$
account for numerical inversion error. Interval arithmetic with
outward rounding also covers normalization and rescaling.
Deterministic numerical verification requires no additional
statistical failure-probability allocation. Verification using stored floating-point features applies to
those values. Guarantees for exact features $\Gamma(x)$ additionally
require bounds on errors in their computation.
For $y\in I$ and $g\in\mathcal G_\kappa(d)$, suppose the computed
prediction $\widetilde y$ satisfies a verified bound
$|g(\Gamma(x))-\widetilde y|\leq\varepsilon$.
A verified upper bound
$\overline e\geq(|y-\widetilde y|+\varepsilon)^2$ then gives
\[
(y-g(\Gamma(x)))^2
\leq\overline\ell
:=\min\{B,\overline e\}
\leq B.
\]
Both arguments of the minimum bound the original
squared loss.
\section{Feature Augmentation and Model Selection}
\label{sec:decisions}
Throughout this section, the optimal risks of candidate classes
and the risks of their fitted predictors are evaluated under
the same joint distribution of $(X,Y)$, with the target $Y$
expressed in the same units for every candidate.
\subsection{Nested Predictor Classes}
\begin{proposition}[Feature augmentation and duplicate features]
\label{prop:refinement}
Fix an integer $s\geq1$, $\kappa\geq0$, the output interval
$I=[y_-,y_+]$, and measurable feature maps
$\Phi_j:\mathcal X\to\mathbb R^{d_j}$,
$j=1,\ldots,s+1$.
For $r\in\{s,s+1\}$, define
$\mathcal U_r=\prod_{j=1}^r\mathbb R^{d_j}$,
$\Gamma_r=(\Phi_1,\ldots,\Phi_r)$, and
$\rho_r(u,v)=\max_{j\leq r}\|u_j-v_j\|_2$.
Let $\mathcal G_\kappa(\rho_r)$ be the class of measurable
decoders $g:\mathcal U_r\to I$ satisfying
$|g(u)-g(v)|\leq\kappa\rho_r(u,v)$ for all
$u,v\in\mathcal U_r$, and set
$\mathcal H_{r,\kappa}
=\{g\circ\Gamma_r:g\in\mathcal G_\kappa(\rho_r)\}$.
Define the clipped affine family
\begin{equation}
\mathcal F_{r,\kappa}
=\left\{x\mapsto\operatorname{clip}_I\!\left(
a+\sum_{j=1}^r\theta_j^\top\Phi_j(x)\right):
\sum_{j=1}^r\|\theta_j\|_2\leq\kappa\right\},
\label{eq:refinement}
\end{equation}
where $a\in\mathbb R$ and $\theta_j\in\mathbb R^{d_j}$. Then $\mathcal F_{r,\kappa}\subseteq\mathcal H_{r,\kappa}$
for $r=s,s+1$, and adding $\Phi_{s+1}$ gives
\begin{equation}
\mathcal F_{s,\kappa}\subseteq\mathcal F_{s+1,\kappa},
\qquad
\mathcal H_{s,\kappa}\subseteq\mathcal H_{s+1,\kappa}.
\label{eq:refinement-inclusions}
\end{equation}
For a fixed joint distribution of $(X,Y)$ with $Y\in I$
almost surely,
\begin{equation}
\begin{aligned}
\inf_{f\in\mathcal F_{s+1,\kappa}}
\mathbb E[(Y-f(X))^2]
&\leq
\inf_{f\in\mathcal F_{s,\kappa}}
\mathbb E[(Y-f(X))^2],\\
\inf_{f\in\mathcal H_{s+1,\kappa}}
\mathbb E[(Y-f(X))^2]
&\leq
\inf_{f\in\mathcal H_{s,\kappa}}
\mathbb E[(Y-f(X))^2].
\end{aligned}
\label{eq:refinement-risk}
\end{equation}
The added group is an exact duplicate of an existing group
if, for some fixed $j\in\{1,\ldots,s\}$,
$d_{s+1}=d_j$ and $\Phi_{s+1}(x)=\Phi_j(x)$ for every
$x\in\mathcal X$.
In this case,
$\mathcal F_{s+1,\kappa}=\mathcal F_{s,\kappa}$ and
$\mathcal H_{s+1,\kappa}=\mathcal H_{s,\kappa}$,
and, for every $x,x'\in\mathcal X$,
\begingroup\postdisplaypenalty=10000
\[
\rho_{s+1}\bigl(\Gamma_{s+1}(x),\Gamma_{s+1}(x')\bigr)
=
\rho_s\bigl(\Gamma_s(x),\Gamma_s(x')\bigr).
\]
Both inequalities in \eqref{eq:refinement-risk} hold with equality.
\endgroup
\end{proposition}
\begin{proof}
For $f\in\mathcal F_{r,\kappa}$, write
$f=q_r\circ\Gamma_r$, where
$q_r(u)=\operatorname{clip}_I(a+\sum_{k=1}^r\theta_k^\top u_k)$
and $\sum_{k=1}^r\|\theta_k\|_2\leq\kappa$.
Clipping is $1$-Lipschitz, so the triangle inequality and
Cauchy-Schwarz give, for all $u,v\in\mathcal U_r$,
\[
|q_r(u)-q_r(v)|
\leq\sum_{k=1}^r\|\theta_k\|_2\|u_k-v_k\|_2
\leq\kappa\rho_r(u,v).
\]
Thus $q_r\in\mathcal G_\kappa(\rho_r)$ and
$f\in\mathcal H_{r,\kappa}$. For $g\in\mathcal G_\kappa(\rho_s)$, define
$g_+(u,w)=g(u)$.
For $u,v\in\mathcal U_s$ and
$w,z\in\mathbb R^{d_{s+1}}$,
\[
|g_+(u,w)-g_+(v,z)|
\leq\kappa\rho_s(u,v)
\leq\kappa\rho_{s+1}((u,w),(v,z)).
\]
Hence $g_+\in\mathcal G_\kappa(\rho_{s+1})$. Moreover,
$g_+(\Gamma_{s+1}(x))=g(\Gamma_s(x))$
for every $x\in\mathcal X$, proving the second inclusion
in \eqref{eq:refinement-inclusions}.
For $g=q_s$, the decoder $g_+$ has the same intercept $a$
and coefficient vectors $(\theta_1,\ldots,\theta_s,0)$.
Their norm sum is
$\sum_{k=1}^s\|\theta_k\|_2+\|0\|_2\leq\kappa$.
Therefore
$g_+\circ\Gamma_{s+1}=q_s\circ\Gamma_s
\in\mathcal F_{s+1,\kappa}$,
proving the first inclusion in \eqref{eq:refinement-inclusions}. Now suppose that $\Phi_{s+1}$ exactly duplicates $\Phi_j$.
For $u,v\in\mathcal U_s$,
\[
\rho_{s+1}((u,u_j),(v,v_j))
=\max\{\rho_s(u,v),\|u_j-v_j\|_2\}
=\rho_s(u,v).
\]
For any $g_+\in\mathcal G_\kappa(\rho_{s+1})$, define
$g(u)=g_+(u,u_j)$. Then
$|g(u)-g(v)|
\leq\kappa\rho_{s+1}((u,u_j),(v,v_j))
=\kappa\rho_s(u,v)$.
Since
$\Gamma_{s+1}(x)=(\Gamma_s(x),\Phi_j(x))$,
we also have
$g(\Gamma_s(x))=g_+(\Gamma_{s+1}(x))$
for every $x\in\mathcal X$.
Thus $\mathcal H_{s+1,\kappa}\subseteq\mathcal H_{s,\kappa}$.
Substituting $u=\Gamma_s(x)$ and $v=\Gamma_s(x')$
into the metric identity proves the stated equality of
distances between represented histories. For any $f_+\in\mathcal F_{s+1,\kappa}$, choose a clipped
affine decoder $g_+$ such that
$f_+=g_+\circ\Gamma_{s+1}$, with intercept $a$ and
coefficients satisfying
$\sum_{k=1}^{s+1}\|\theta_k\|_2\leq\kappa$.
The decoder $g(u)=g_+(u,u_j)$ has the explicit form
\[
g(u)=\operatorname{clip}_I\!\left(
a+\sum_{\substack{1\leq k\leq s\\k\ne j}}
\theta_k^\top u_k
+(\theta_j+\theta_{s+1})^\top u_j\right).
\]
Its coefficients satisfy
\[
\sum_{\substack{1\leq k\leq s\\k\ne j}}\|\theta_k\|_2
+\|\theta_j+\theta_{s+1}\|_2
\leq\sum_{k=1}^{s+1}\|\theta_k\|_2
\leq\kappa.
\]
Consequently,
$g\circ\Gamma_s=f_+\in\mathcal F_{s,\kappa}$,
so $\mathcal F_{s+1,\kappa}\subseteq\mathcal F_{s,\kappa}$.
Together with \eqref{eq:refinement-inclusions}, these two
reverse inclusions establish the stated class equalities. Each left-hand infimum in \eqref{eq:refinement-risk} is
taken over a class containing the corresponding class on
the right, which proves both risk inequalities.
For an exact duplicate, the corresponding classes are equal,
so their infimum risks are equal.
\end{proof}
Successively adding feature groups while keeping the existing
feature maps, $I$, and $\kappa$ fixed yields nested sequences
of clipped affine and Lipschitz predictor classes.

\subsection{Risk-Threshold Decisions and Minimum Cost}

\emph{Candidates and their costs.}
A candidate $m$ is a pair $(\Gamma_m,\mathcal G_m)$,
where $\Gamma_m:\mathcal X\to\mathcal U_m$ is a measurable
representation of histories $x\in\mathcal X$, and
$\mathcal G_m$ is a specified class of measurable decoders
$g:\mathcal U_m\to I$, with prediction interval
$I=[y_-,y_+]$.
Its predictor class is
$\{g\circ\Gamma_m:g\in\mathcal G_m\}$.
A common cost measure assigns a finite nonnegative cost
$c_m$ to candidate $m$, with the same assigned cost for
every decoder choice $g\in\mathcal G_m$ within that candidate.
The candidates, cost measure, and costs are fixed before
calibration.
If cost is the number of feature coordinates, then
$c_m=p_m$ for $\Gamma_m:\mathcal X\to\mathbb R^{p_m}$.
\begin{corollary}[Risk-threshold decisions and minimum cost]
\label{cor:decisions}
Let $m=1,\ldots,M$ index a finite collection of
candidates $(\Gamma_m,\mathcal G_m)$ with costs
$c_m$ and measurable fitted decoders
$\widehat g_m:\mathcal U_m\to I$.
Define
\[
\mathcal A
=\{m:\widehat g_m\in\mathcal G_m\}.
\]
Assume the conditions of Theorem~\ref{thm:certificate}
for these candidates, fitted decoders, and confidence
bounds $L_m,U_m$, with total allocated failure probability
at most $\delta\in(0,1)$. All candidates predict the same target $Y$, expressed in
the same units, under a common evaluation distribution
of $(X,Y)$. Define
\[
R_m(g)=\mathbb E[(Y-g(\Gamma_m(X)))^2],
\qquad
R_m^*=\inf_{g\in\mathcal G_m}R_m(g).
\]
Let $\tau\geq0$ be a population mean squared error threshold
fixed before calibration. With $\min\varnothing=+\infty$, define the minimum assigned cost of meeting the threshold within this candidate
collection by
\[
c^*=\min\{c_m:\exists g\in\mathcal G_m
\text{ with }R_m(g)\leq\tau\}.
\]
By Theorem~\ref{thm:certificate}, conditional
on the development data, with probability at
least $1-\delta$, the inequalities
$L_m\leq R_m^*$ and
$R_m(\widehat g_m)\leq U_m$
hold simultaneously for all candidates $m$. On this event, for every candidate $m$,
\begin{equation}
\begin{aligned}
m\in\mathcal A\ \text{and}\ U_m\leq\tau
&\ \Longrightarrow\
R_m^*\leq R_m(\widehat g_m)\leq\tau,\\
L_m>\tau
&\ \Longrightarrow\
R_m(g)>\tau
\quad\text{for every }g\in\mathcal G_m.
\end{aligned}
\label{eq:threshold-decisions}
\end{equation}
On the same event, the minimum cost satisfies
\begin{equation}
c_-:=\min_{m:L_m\leq\tau}c_m
\ \leq\ c^*\ \leq\
c_+:=\min_{\substack{m\in\mathcal A\\U_m\leq\tau}}c_m.
\label{eq:cost}
\end{equation}
If $c_+<\infty$, any index
\[
\widehat m\in
\operatorname*{arg\,min}_{\substack{
m\in\mathcal A\\U_m\leq\tau}}c_m
\]
satisfies
$R_{\widehat m}(\widehat g_{\widehat m})\leq\tau$
and $c_{\widehat m}=c_+$. If $c_-=c_+<\infty$, then $c_{\widehat m}=c^*$. If $c_-=+\infty$, no candidate class contains a decoder
with population mean squared error at most $\tau$.
\end{corollary}

\begin{proof}
On the simultaneous event
\eqref{eq:separate-confidence}, we have
$L_m\leq R_m^*$ and
$R_m(\widehat g_m)\leq U_m$ for every
candidate $m$. Since
$R_m(g)\geq R_m^*$ for every
$g\in\mathcal G_m$, the condition
$L_m>\tau$ excludes every decoder in
$\mathcal G_m$. If $m\in\mathcal A$ and $U_m\leq\tau$,
then $\widehat g_m\in\mathcal G_m$ and
$R_m(\widehat g_m)\leq\tau$.
Consequently,
\[
\begin{aligned}
\{m\in\mathcal A:U_m\leq\tau\}
&\subseteq
\{m:\exists g\in\mathcal G_m,\
R_m(g)\leq\tau\} 
&\subseteq
\{m:L_m\leq\tau\}.
\end{aligned}
\]
The second inclusion holds because
$R_m(g)\leq\tau$ implies
$L_m\leq R_m^*\leq R_m(g)\leq\tau$.
Taking the minimum assigned cost over
these three sets establishes
\eqref{eq:cost}. If $c_+<\infty$, the finite, nonempty
set $\{m\in\mathcal A:U_m\leq\tau\}$
has a minimizing index $\widehat m$.
This index satisfies
$c_{\widehat m}=c_+$ and
$R_{\widehat m}(\widehat g_{\widehat m})
\leq U_{\widehat m}\leq\tau$.
If $c_-=c_+<\infty$, then
\eqref{eq:cost} gives
$c_{\widehat m}=c^*$. If $c_-=+\infty$, then
$L_m>\tau$ for every candidate $m$.
Hence $R_m(g)>\tau$ for every
$g\in\mathcal G_m$ and every $m$.
Theorem~\ref{thm:certificate} gives
conditional probability at least
$1-\delta$ for the simultaneous
event used throughout the proof.
\end{proof}
\subsection{Risk Comparisons and Sample Size Requirements}
\begin{proposition}[Risk comparison and necessary sample size] \label{prop:risk-comparison}
Assume the conditions of Theorem~\ref{thm:certificate}.
All candidates predict the same target $Y$, expressed in
the same units, under a common evaluation distribution
of $(X,Y)$. Let
$R_m(g)=\mathbb E[(Y-g(\Gamma_m(X)))^2]$
denote the population mean squared error for candidate $m$. On the simultaneous event
\eqref{eq:separate-confidence}, which has conditional probability at least $1-\delta$
given the development data, every pair of candidates $c,r$ with $L_c>U_r$ satisfies
\begin{equation}
R_c(g)-R_r(\widehat g_r)
\geq L_c-U_r>0
\qquad\text{for every }g\in\mathcal G_c.
\label{eq:risk-comparison}
\end{equation}
On the same event, for a population mean squared error
threshold $\tau\geq0$, the condition $U_r\leq\tau<L_c$
implies
$R_r(\widehat g_r)\leq\tau<R_c(g)$
for every $g\in\mathcal G_c$. For the KL upper bound $U_r$ in \eqref{eq:upper}, let
$n\geq1$ be the number of independent calibration
observations, $B>0$ the squared-loss range bound, and
$\alpha=\alpha_r^{\rm up}\in(0,1)$.
For every calibration data set,
\begin{equation}
U_r\geq B(1-\alpha^{1/n})>0.
\label{eq:floor}
\end{equation}
If all numerical loss upper bounds
$\overline\ell_{r,i}$ are zero, exact evaluation of
\eqref{eq:upper} gives $U_r=B(1-\alpha^{1/n})$.
For $0<\tau<B$, the condition $U_r\leq\tau$ requires
\begin{equation}
n\geq
\left\lceil
\frac{\log(1/\alpha)}{-\log(1-\tau/B)}
\right\rceil.
\label{eq:kl-sample-size}
\end{equation}
\end{proposition}
\begin{proof}
On the simultaneous event in \eqref{eq:separate-confidence}, $R_c(g)\geq L_c$ for every $g\in\mathcal G_c$
and $R_r(\widehat g_r)\leq U_r$.
Subtracting gives \eqref{eq:risk-comparison} when
$L_c>U_r$; the same inequalities give the threshold
comparison when $U_r\leq\tau<L_c$. Since $\operatorname{kl}(0\Vert p)=-\log(1-p)$,
$k_+(0,\log(1/\alpha)/n)=1-\alpha^{1/n}$.
The input $(nB)^{-1}\sum_{i=1}^n\overline\ell_{r,i}$
in \eqref{eq:upper} is nonnegative, so monotonicity of
$k_+$ in this input gives \eqref{eq:floor}, with
$U_r=B(1-\alpha^{1/n})$ under exact evaluation when
all $\overline\ell_{r,i}=0$.
For $0<\tau<B$, \eqref{eq:floor} and $U_r\leq\tau$
imply $(1-\tau/B)^n\leq\alpha$.
Taking logarithms and using that $n$ is an integer
gives \eqref{eq:kl-sample-size}.
\end{proof}
For a block-based lower bound $L_c=L^{\rm blk}_{c,b}$,
Appendix~\ref{app:resolution} gives sufficient conditions
for $L_c>U_r$ and $U_r\leq\tau<L_c$.

%% file: sections/representations.tex
\section{Feature Representations and Evaluation Protocols}
\label{sec:representations}
The experiments compare representations based on Kernel Affine
Hull Machines (KAHMs)
\citep{kumar2024kahm,kumar2025collaborative,kumar2026limits}
with raw-history and Nystr\"om representations.
KAHM features describe how well local models reconstruct
a preprocessed history.
Each forecasting task specifies the history length,
forecast horizon, and target normalization.
Candidates within a comparison use the same target definition
and evaluation law or assessment observations.

\subsection{Features from Local Reconstruction Models}

For a history $x$, let $z(x)\in\mathbb R^{d_x}$ be its
preprocessed vector, with preprocessing parameters fitted
using development inputs.
The affinity representation contains $s\geq1$ feature groups.
Group $j$ uses $C_j\geq1$ reference sets constructed from
development inputs.
Reference set $c$ is used to fit $K_{jc}\geq1$ local KAHMs.
Let $\widehat z_{jck}(x)\in\mathbb R^{d_x}$ be the
reconstruction returned by model $k=1,\ldots,K_{jc}$.
Define the clipped cosine similarity $\gamma_{jck}$,
transformed Euclidean reconstruction error $E_{jck}$,
and normalized angular distance $A_{jck}$ by
\begin{align*}
\gamma_{jck}(x)
&=
\begin{cases}
\operatorname{clip}_{[-1,1]}\!\left(
\dfrac{z(x)^\top\widehat z_{jck}(x)}
{\|z(x)\|_2\|\widehat z_{jck}(x)\|_2}
\right),
& \|z(x)\|_2\|\widehat z_{jck}(x)\|_2>0,\\
0,&\text{otherwise},
\end{cases}\\
E_{jck}(x)
&=1-\exp\{-\|z(x)-\widehat z_{jck}(x)\|_2\},
\qquad
A_{jck}(x)=\pi^{-1}\arccos\gamma_{jck}(x).
\end{align*}
The zero-vector convention gives $A_{jck}(x)=1/2$
when either vector is zero.
For fixed $\eta>0$ and $\omega>0$, define the reconstruction
score and normalized affinity by
\begin{align*}
T_{jc}(x)
&=\min_{1\leq k\leq K_{jc}}
\sqrt{\frac{E_{jck}(x)^2+A_{jck}(x)^2}{2}}, \quad
\Phi_{jc}(x)
&=\frac{(1-T_{jc}(x)+\eta)^\omega}
{\sum_{r=1}^{C_j}(1-T_{jr}(x)+\eta)^\omega}.
\end{align*}
Each reference set contributes one affinity coordinate.
In exact arithmetic, these coordinates are positive and
sum to one within each group.
Writing $\Phi_j=(\Phi_{j1},\ldots,\Phi_{jC_j})$, the map
$\Gamma_s=(\Phi_1,\ldots,\Phi_s)$ has
$\sum_{j=1}^s C_j$ coordinates. 
The feature-augmentation experiment uses two
groups containing $C_1=2$ and $C_2=4$ local KAHM
models, respectively, with one model fitted to
each reference set. It compares the two-coordinate
representation $\Gamma_1=\Phi_1$ with the
six-coordinate representation
$\Gamma_2=(\Phi_1,\Phi_2)$. In the finite-state demonstrator, we denote the
six-coordinate normalized-affinity representation
by MEM. It uses the same group sizes
$C_1=2$ and $C_2=4$, but its local KAHM models
are fitted separately using the demonstrator's
training data. The LOCAL representation augments MEM with
affinity-weighted local projection coordinates.
For each local model $c$ in group $j$, let
$\mu_{jc}$ denote the mean of its reference
vectors, and let the columns of $P_{jc}$
contain the principal component directions
retained by that model. LOCAL appends $\Phi_{jc}(x)P_{jc}^{\top}(z(x)-\mu_{jc})$ for each of the six local models. Each selected
model retains two principal component directions,
so LOCAL contains $6+6\times2=18$ coordinates. Appendix~\ref{app:feature-implementation}
describes the numerical conventions used in
the feature-augmentation experiment.

\subsection{Feature Geometry and Lipschitz Constraints}
The feature-augmentation and PTB-XL studies use the maximum
groupwise Euclidean metric $\rho_s(u,v)=\max_{j\leq s}\|u_j-v_j\|_2$. The demonstrator and Bike Sharing studies use Euclidean
distance on their declared representation coordinates.
In the demonstrator, these are the coordinates before the
readout's internal scaling and Nystr\"om transformation.
For Bike Sharing, each map is multiplied by the reciprocal
median positive Euclidean distance between feature vectors
of 32 training histories.
The scale is one if no positive distance exists.
The stored feature coordinates include this normalization. For the exact affinity formula, let
$\Delta_C=\{p\in[0,1]^C:\sum_c p_c=1\}$ and
$\mathcal K_s=\prod_{j=1}^s\Delta_{C_j}$.
Then $\Gamma_s$ takes values in $\mathcal K_s$, which is compact
and has diameter at most $\sqrt2$ under $\rho_s$. Each demonstrator representation has finite support under
its evaluation law, so its stored feature domain is compact
under Euclidean distance. An $I$-valued $\kappa$-Lipschitz decoder on either domain extends
to the ambient feature space with the same Lipschitz bound
and the same predictions on that domain.
The minimum risk under the specified evaluation law is
therefore the same for the compact and ambient domains.
On these compact domains, for a fixed measurable representation,
fixed evaluation law, and fixed $\kappa$,
Theorem~\ref{thm:hierarchy} gives
$\mathcal W_b\uparrow R_\kappa^*$ as $b\to\infty$ under iid
sampling with $Y\in I$ almost surely. Changing input preprocessing can alter the fitted local models
and the resulting features.
For an already fixed feature map, both metrics satisfy
$d(au,av)=a\,d(u,v)$ for $a>0$.
The transformation $g_a(v)=g(v/a)$ therefore preserves all
predictions when the features are multiplied by $a$ and the
Lipschitz bound is changed from $\kappa$ to $\kappa/a$. Holding $\kappa$ fixed while changing feature scale can change
the admissible class and its minimum risk. 

\subsection{Evaluation Protocols}
In both controlled experiments, development data fix the
representations, decoder classes, and fitted predictors before
independent iid calibration observations are used.
Lower confidence bounds concern minimum population risk over
the specified classes.
An upper confidence bound on fitted-predictor risk also bounds
the class minimum when the fitted predictor is admissible.
The demonstrator checks admissibility on its finite support;
the feature-augmentation experiment uses the clipped affine
family in \eqref{eq:refinement}, subject to the numerical
conditions stated in
Section~\ref{sec:feature-augmentation-experiment}. In feature augmentation, retaining the original feature group
with the same $I$ and $\kappa$ gives nested Lipschitz and clipped
affine predictor classes
(Proposition~\ref{prop:refinement}).
The demonstrator also has nested classes from MEM to LOCAL,
and from short history to full history.
Within each pair, $I$ and $\kappa$ agree.
A decoder can therefore ignore the added coordinates because
coordinate projection does not increase Euclidean distance.
These inclusions concern admissible predictor classes and
do not order the risks of separately fitted predictors.
The PTB-XL and Bike Sharing maps are not assumed to define
nested predictor classes. PTB-XL and Bike Sharing instead evaluate minimum empirical risk on fixed assessment observations.
Their deterministic bounds use \eqref{eq:cohort}.
This optimization constrains predictions to $I$ while allowing
assessment targets outside $I$, as occurs for Bike Sharing.
The population-risk confidence bounds used here assume
$Y\in I$ almost surely.

%% file: sections/evidence.tex
\section{Empirical Evaluation}
\label{sec:evidence}
\subsection{Block Hierarchy and Verified Class-Risk Decisions}
\label{sec:demonstrator}
\emph{Forecasting task.}
We use a finite-state forecasting experiment with a
known evaluation distribution to compare calibrated
risk bounds with population class risks computed
independently of calibration. Let
$v=(0,1/4,3/4,1)$ denote the observed values
associated with states $0,1,2,3$. Each episode
begins by drawing $(S_{-1},S_0)$ uniformly from
$\{0,1,2,3\}^2$. The next state $S_1$ is then
drawn according to
\[
\Pr(S_1=k\mid S_{-1}=i,S_0=j)
=\frac45\mathbf 1\{k=(i+j)\bmod4\}+\frac1{20}.
\]
The predictor receives the two-observation history
$X=(v_{S_{-1}},v_{S_0})$, ordered from previous
to current observation, and predicts $Y=v_{S_1}$.
The evaluation distribution contains 16 possible
histories and 64 history-target outcomes.
Calibration observations consist of independent
episodes drawn from this distribution. Targets
and predictions lie in $I=[0,1]$, and the
prescribed MSE threshold is $\tau=0.1$.

\emph{Representations and fitted predictors.}
We compare five representation-decoder classes using
Euclidean feature distances. RAW consists of the two
history coordinates standardized using training means
and standard deviations. MEM contains six KAHM
reconstruction affinities, normalized separately
within groups of two and four local models. 
LOCAL retains the six MEM affinity coordinates
and appends two affinity-weighted local projection
coordinates from each of the six KAHM models,
giving 18 coordinates. MEM and LOCAL use the
same fitted KAHM models, each with a
two-dimensional local subspace obtained by
principal component analysis (PCA). The affinity
exponent is $\omega=16$, and the offset is
$\eta=10^{-6}$. The RAW, MEM, and LOCAL
classes use $\kappa=2$. The short-history control
uses $v_{S_0}$, whereas the full-history control uses
$(v_{S_{-1}},v_{S_0})$. Both controls use their
original observation values with $\kappa=4$. The representations and fitted predictors are
developed using 4096 training episodes and 1024
separate validation episodes. For RAW, MEM, and
LOCAL, the fitted predictors use ridge regression
on the representation coordinates and Gaussian
Nystr\"om features, with predictions clipped to
$[0,1]$. The short-history control predicts $Y$ using only
the current observation $v_{S_0}$. Its prediction
is the average target value over all training
episodes with the same value of $v_{S_0}$. 
The full-history control predicts $Y$ using both
the previous observation $v_{S_{-1}}$ and the
current observation $v_{S_0}$. Its prediction
is the average target value over all training
episodes with the same ordered pair
$(v_{S_{-1}},v_{S_0})$. If the current observation $v_{S_0}$ does not occur
in the training data, the short-history control
predicts the mean target value across all training
episodes. Similarly, if the ordered pair
$(v_{S_{-1}},v_{S_0})$ does not occur in the
training data, the full-history control uses
the same overall training-target mean.
 All representations and fitted prediction
rules are fixed before calibration. The Lipschitz constraints defining each decoder
class apply to the declared representation
coordinates, before any internal scaling or
Nystr\"om transformation used by the fitted
predictor. For the numerical assessment, we treat
the stored feature vectors and fitted predictions
on all 16 possible histories as exact finite
tables. A fitted prediction table is admissible
if its predictions lie in $I$ and satisfy the
prescribed Lipschitz constraints for every pair
of histories. If these conditions hold, there exists a decoder
defined on the entire feature space that satisfies
the prescribed Lipschitz constraint and reproduces
the fitted predictions for all 16 possible histories.

\emph{Calibration.}
We generate a calibration dataset containing 16,384
independent and identically distributed (iid)
observations. For each calibration sample size
$n\in\{1024,4096,16384\}$, we use the first $n$
observations from this dataset. For each decoder
class and calibration sample size, we partition
the observations into consecutive, nonoverlapping
blocks of each prespecified size
$b\in\mathcal B=\{2,8,16,32,64\}$.
Each partition contains $n/b$ blocks. We compute verified lower bounds on the block
optimization problem in \eqref{eq:block} using
the dual formulation of
Proposition~\ref{prop:dual}. For each decoder
class $m$ and block, we first aggregate observations
with identical feature vectors. We then use cyclic
coordinate ascent with soft-thresholding to obtain
numerical dual multipliers. The algorithm performs
at most 100 coordinate sweeps and selects pairwise
Lipschitz constraints satisfying
$\kappa_m d_{ij}^{+}<1$, where $d_{ij}^{+}$ is a
verified upper bound on the Euclidean distance
between the corresponding feature vectors.
We evaluate the dual objective in \eqref{eq:dual}
using exact rational arithmetic. The resulting
nonnegative lower bound on the minimum total
squared error, divided by $b-1$, defines
$\underline T_{m,b,k}$ for block $k$, satisfying
\[
0\leq\underline T_{m,b,k}\leq T_{m,b,k},
\]
where $T_{m,b,k}$ is the exact normalized block
optimum. Proposition~\ref{prop:dual} guarantees
the validity of these lower bounds even if
coordinate ascent does not converge. We use the verified block lower bounds in two
calibration methods. Direct block calibration
applies \eqref{eq:Lblock} to obtain
$L^{\rm blk}_{m,b}$ from the $n/b$ independent
blocks. Empirical-risk calibration combines the
block lower bounds to obtain
\[
d_{m,n,b}
=\frac{b-1}{n}\sum_{k=1}^{n/b}\underline T_{m,b,k}
\leq a_{m,n},
\]
where $a_{m,n}$ is the minimum empirical MSE
over all $n$ calibration observations. The
inequality follows from \eqref{eq:normalization}.
Applying \eqref{eq:Lemp} to $d_{m,n,b}$ gives
$L^{\rm emp}_{m,b}$, a lower confidence bound
on optimal population class risk. We additionally compute a verified lower bound
on the minimum empirical MSE by considering all
$n$ calibration observations jointly in a single
optimization problem, using the dual procedure
of Proposition~\ref{prop:dual}.
Dividing the resulting lower bound on minimum
total squared error by $n$ gives a lower bound
on $a_{m,n}$. Applying \eqref{eq:Lemp} to this
bound yields $L_m^{\rm full}$, a lower confidence
bound on optimal population class risk. For each decoder class, the reported lower
confidence bound is
\[
L_m=\max\left\{
\max_{b\in\mathcal B}L^{\rm blk}_{m,b},\;
\max_{b\in\mathcal B}L^{\rm emp}_{m,b},\;
L^{\rm full}_m\right\}.
\]
The upper confidence bound $U_{{\rm fit},m}$
on fitted-predictor risk is computed using
\eqref{eq:upper}. For each of the five classes and three calibration
sample sizes, we compute five block-calibration
bounds, five empirical-risk calibration bounds,
one full-sample bound, and one fitted-risk upper
bound. This gives $5\times3\times12=180$
confidence bounds: 165 lower bounds on optimal
population class risk and 15 upper bounds on
fitted-predictor risk. We allocate failure
probability $\alpha=1/3600$ to each bound.
Conditional on the fixed development data
and under the stated iid sampling assumptions,
the union bound guarantees simultaneous
coverage of at least $0.95$.

\begin{figure}[tbp]
\centering
\includegraphics[width=0.94\linewidth]{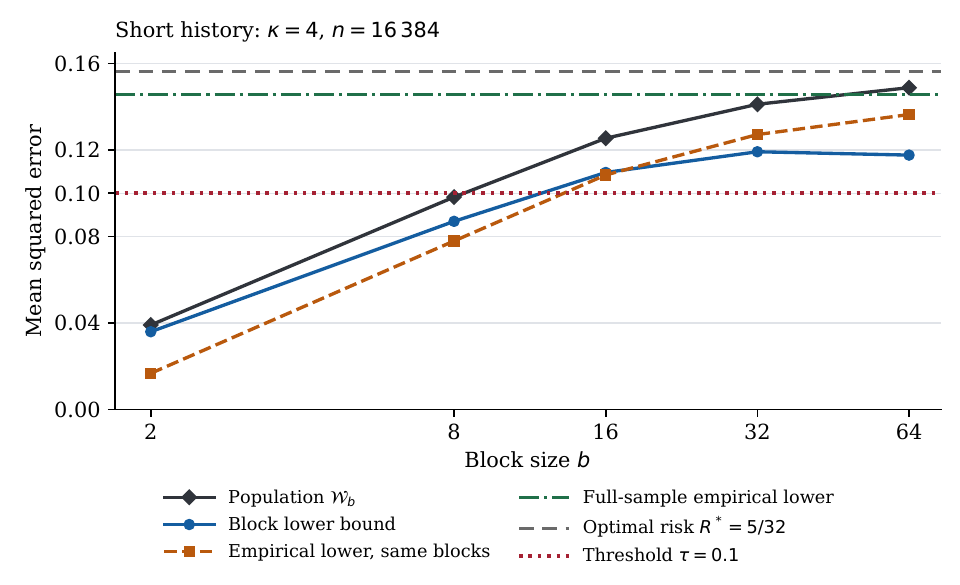}
\caption{Short-history class at $\kappa=4$ and $n=16384$.
The expected block minimum $\mathcal W_b$ increases toward
the optimal population risk $5/32$. The block and empirical-risk
lower confidence bounds use the same block optimization results;
neither method gives the larger lower confidence bound at
every displayed block size. The full-sample empirical-risk lower bound uses
all calibration observations jointly. The horizontal threshold
is $\tau=0.1$. All confidence bounds belong to the same
prespecified family with failure allocation $1/3600$ per bound.}
\label{fig:demonstrator-hierarchy}
\end{figure}

\emph{Effect of block size.}
We use the short-history class to examine how the
expected block scores in Theorem~\ref{thm:hierarchy}
and their calibrated lower confidence bounds change
with block size. Under the evaluation distribution,
$Y$ is uniformly distributed over the four observation
values and is independent of the current observation
$v_{S_0}$. The constant predictor $1/2$ therefore
attains the optimal population MSE of the short-history
class, $R_\kappa^*=\operatorname{Var}(Y)=5/32$. Distinct current observation values are separated
by at least $1/4$. Consequently, the Lipschitz
constraint with $\kappa=4$ permits any predictions
in $[0,1]$ at these distinct values. For a block
of $b$ observations, the minimum total squared
error is therefore obtained by predicting the
sample mean target for each distinct current
observation value. Let $J_b$ denote the number
of distinct current observation values in the
block. Conditional on these values, the expected
minimum total squared error is $(b-J_b)5/32$.
Since $\mathbb E[J_b]=4[1-(3/4)^b]$, the expected
normalized block score is
\begin{equation}
\mathcal W_b
=\frac5{32}\frac{b-4+4(3/4)^b}{b-1}.
\label{eq:demonstrator-hierarchy}
\end{equation}
In particular, $\mathcal W_2=0.0390625$ and
$\mathcal W_8\approx0.09822$ are below the
prescribed threshold $\tau=0.1$, whereas
$\mathcal W_{16}\approx0.12542$ exceeds it. Figure~\ref{fig:demonstrator-hierarchy} compares
these expected block scores with the lower
confidence bounds obtained from the calibration
dataset of size $n=16384$. The lower confidence
bound from direct block calibration increases
from approximately $0.03595$ at $b=2$ to
$0.10960$ at $b=16$. Block sizes 16, 32, and 64
therefore establish that the short-history class
cannot meet the prescribed population MSE threshold.
However, the lower confidence bound decreases
from approximately $0.11918$ at $b=32$ to
$0.11763$ at $b=64$. Although the expected
block score $\mathcal W_b$ increases with $b$,
larger blocks provide fewer independent block
scores for calibration. Consequently, the
finite-sample lower confidence bounds need
not increase with block size.
For the short-history class, the lower confidence
bound obtained by optimizing all calibration
observations jointly is approximately $0.11368$
at $n=1024$, exceeding the prescribed MSE
threshold $\tau=0.1$. At $n=16384$, this bound
increases to approximately $0.14579$, exceeding
the lower confidence bounds obtained from
separate block optimizations.

\emph{Class decisions and admissibility.}
Table~\ref{tab:demonstrator-decisions} reports the
class-risk decisions for the five representation--decoder
classes at calibration sample size $n=4096$ and
MSE threshold $\tau=0.1$. The lower confidence
bounds for the MEM and short-history classes exceed
the threshold, establishing their exclusion at the
prescribed confidence level. For RAW, LOCAL, and
full history, the fitted predictors satisfy the
prescribed Lipschitz constraints, and their upper
confidence bounds on population MSE are below
the threshold. These results establish that
the three corresponding decoder classes meet
the MSE requirement.
\begin{table}[tbp]
\centering\small
\begin{tabular}{@{}lcrrcl@{}}
\toprule
Representation & $\kappa$ & $L_m$ & $U_{{\rm fit},m}$ &
\shortstack{Fitted table\\admissible} & Class decision\\
\midrule
RAW           & 2 & 0.042810 & 0.072273 & Yes & Requirement met\\
MEM           & 2 & 0.114702 & 0.072273 & No  & Excluded\\
LOCAL         & 2 & 0.042810 & 0.072273 & Yes & Requirement met\\
Short history & 4 & 0.134763 & 0.180789 & Yes & Excluded\\
Full history  & 4 & 0.042810 & 0.072273 & Yes & Requirement met\\
\bottomrule
\end{tabular}
\caption{Class-risk decisions for $n=4096$ and $\tau=0.1$.
$L_m$ bounds optimal class risk from below, whereas
$U_{{\rm fit},m}$ bounds the risk of the fitted prediction table
from above. The MEM upper bound is not a class-risk upper bound
because its fitted table is inadmissible. Displayed lower bounds
are rounded downward and upper bounds upward to six decimal
places; the decisions use the full-precision stored endpoints.}
\label{tab:demonstrator-decisions}
\end{table}
Both full-sample
and block-based empirical-risk calibration exclude
the MEM and short-history classes, whereas direct
block calibration does not exclude either class.
The archived numerical results show that full-sample
empirical-risk calibration produces the largest
lower confidence bound in all 15 combinations of
five decoder classes and three calibration sample
sizes. At $n=1024$, the MEM class remains unresolved:
its lower confidence bound is approximately
$0.09582$, below the prescribed threshold
$\tau=0.1$, while its fitted predictor does not
satisfy the prescribed Lipschitz constraint
and therefore cannot establish class feasibility. The fitted MEM predictor has population MSE
approximately $0.05644$, but violates the
prescribed Lipschitz constraint with $\kappa=2$.
Its low prediction error therefore does not
contradict the exclusion of the MEM decoder class. Using the known evaluation distribution, we
compute a verified lower bound on the optimal
population MSE of the MEM class from the dual
optimization problem and an upper bound from
an admissible decoder. These bounds place the
optimal population MSE between $0.13353472$
and $0.13354831$. This calculation does not
use the calibration observations and provides
a reference for evaluating the calibrated
confidence bounds. The optimal population MSE
of the RAW, LOCAL, and full-history classes
is exactly $9/160=0.05625$.

\subsection{Feature Augmentation and Calibration Sample Size} \label{sec:feature-augmentation-experiment}
This experiment examines how feature augmentation
and calibration sample size affect risk-threshold
decisions when the observed history contains
predictive information about the target and when
the target is independent of the history.
We repeat representation learning, predictor
fitting, and calibration twelve times for each
of two data-generating distributions. In both settings, the observed history is
$X=(X_1,X_2)$, where
$X_j=S_j(0.9+V_j)$ for $j=1,2$.
The variables $S_1,S_2,V_1,V_2$ are mutually
independent, with
$\Pr(S_j=1)=\Pr(S_j=-1)=1/2$ and
$V_j\sim\operatorname{Unif}[-0.1,0.1]$.
Let $X'$ be an independent copy of $X$, and let
$E\sim\operatorname{Unif}[-0.05,0.05]$ be
independent of both. In the history-dependent setting, the target is
$Y=0.8X_1X_2+E$. In the independent-target
setting, the target is $Y=0.8X'_1X'_2+E$.
Predictors receive only $X$ in both settings.
The two settings have the same marginal target
distribution but differ in whether $Y$ depends
on the observed history. Targets and clipped
predictions lie in $I=[-0.85,0.85]$, giving
the squared-loss bound $B=2.89$.
The prescribed MSE threshold is $\tau=0.1$.

\emph{Nested representations.}
We compare two KAHM affinity representations.
The first representation, $\Gamma_1=\Phi_1$,
contains two affinity coordinates. The augmented
representation, $\Gamma_2=(\Phi_1,\Phi_2)$,
retains these two coordinates and adds four
affinity coordinates, giving six coordinates
in total (Section~\ref{sec:representations}). For $m\in\{1,2\}$, the corresponding decoder
class consists of all measurable functions
into $I$ satisfying a Lipschitz constraint
with constant $\kappa=2$ under the maximum
groupwise Euclidean metric $d_m(u,v)=\max_{1\leq j\leq m}\|u_j-v_j\|_2$. Because $\Gamma_2$ retains the complete
representation $\Gamma_1$, every admissible
decoder using $\Gamma_1$ can also be used
with $\Gamma_2$ by ignoring the additional
feature group $\Phi_2$. Consequently, the
optimal population risks satisfy
$R_2^*\leq R_1^*$
(Proposition~\ref{prop:refinement}). In the independent-target setting, $Y$ is
independent of $X$ and has mean zero.
Therefore, the constant zero predictor
achieves the minimum population MSE for
both decoder classes. Each repetition uses separately generated,
disjoint datasets for representation learning,
decoder fitting, validation, and calibration.
We fit predictors from the clipped affine
family defined in \eqref{eq:refinement}.
The representations and fitted predictors
are fixed before calibration.
Appendix~\ref{app:protocols} specifies the
sample splits and hyperparameter selection.

\emph{Calibration and comparison.}
We compare direct block calibration and
empirical-risk calibration using the same
calibration observations and verified block
optimization results. For each of the two
data-generating settings, we perform 12
independent experimental runs. Each run uses
newly generated datasets for representation
learning, predictor fitting, validation, and
calibration. The calibration dataset contains
4096 iid observations. We use the first $n$
observations for each prespecified calibration
sample size $n\in\{256,1024,4096\}$, with
$n=1024$ designated as primary. For each
sample size, we partition the observations
into consecutive, nonoverlapping blocks of
sizes $b\in\mathcal B=\{2,4,8,16,32\}$. Direct block calibration applies \eqref{eq:Lblock}
to the verified lower bounds on normalized
block scores and reports $L_m^{\rm blk}
=\max_{b\in\mathcal B}L_{m,b}^{\rm blk}$. Empirical-risk calibration combines the verified
block lower bounds to obtain
\[
d_{m,n}
=\max_{b\in\mathcal B}
\frac{b-1}{n}\sum_{k=1}^{n/b}
\underline T_{m,b,k}
\leq a_{m,n},
\]
where $a_{m,n}$ is the minimum empirical MSE
over all $n$ calibration observations.
The inequality follows from
\eqref{eq:normalization}. Applying
\eqref{eq:Lemp} to $d_{m,n}$ yields
$L_m^{\rm emp}$, a lower confidence bound
on optimal population class risk.
For both methods, we compute the fitted-predictor
risk upper bounds $U_{{\rm fit},m}$ using
\eqref{eq:upper}. For each independent experimental run, we
evaluate two decoder classes at three calibration
sample sizes. Each class and sample size yields
five direct block lower confidence bounds,
one empirical-risk lower confidence bound,
and one fitted-predictor risk upper confidence
bound. Across the 24 experimental runs, this
gives $24\times2\times3\times7=1008$
confidence bounds: 864 lower bounds on
optimal population class risk and 144 upper
bounds on fitted-predictor risk.
We allocate failure probability
$\alpha=0.05/1008$ to each bound. Under the sampling assumptions and numerical
requirements of Theorem~\ref{thm:certificate}, the union bound guarantees simultaneous coverage of at least $0.95$ across the experiment. The numerical verification is incomplete (Section~\ref{sec:discussion}).
\begin{table}[tbp]
\centering\small
\begin{tabular}{@{}llrr@{}}
\toprule
Target setting & Calibration size &
\shortstack{Block\\calibration} &
\shortstack{Empirical-risk\\calibration}\\
\midrule
History-dependent & 256 (secondary)  & 0  & 0\\
History-dependent & 1024 (primary)   & 12 & 12\\
History-dependent & 4096 (secondary) & 12 & 12\\
\addlinespace
Independent target & 256 (secondary)  & 1  & 12\\
Independent target & 1024 (primary)   & 12 & 12\\
Independent target & 4096 (secondary) & 12 & 12\\
\bottomrule
\end{tabular}
\caption{Numbers of conclusive comparisons
out of 12 independent experimental runs per
target setting. For method $q\in\{\mathrm{blk},\mathrm{emp}\}$, a run is
counted if $U_{{\rm fit},2}\leq\tau<L_1^q$
in the history-dependent setting, or if
$L_1^q>\tau$ and $L_2^q>\tau$ in the
independent-target setting. Both methods use
the same fitted-predictor risk upper bounds,
and $\tau=0.1$.}
\label{tab:controlled}
\end{table}

\emph{Results and interpretation.}
Table~\ref{tab:controlled} summarizes the threshold
comparisons. The class decisions below are conditional
on the numerical requirements and, for class feasibility,
fitted-decoder admissibility. At the primary calibration sample size
$n=1024$, both calibration methods produce
conclusive decisions in all 12 runs under each
setting. In the history-dependent setting, the
two-coordinate decoder class is excluded, while
an admissible fitted predictor using the
six-coordinate representation meets the MSE
threshold. In the independent-target setting,
both decoder classes are excluded. At $n=256$, empirical-risk calibration excludes
both decoder classes in all 12 runs under the
independent-target setting, whereas direct block
calibration does so in only one run. In the history-dependent setting, neither
calibration method produces a conclusive
comparison at $n=256$. The reason is that
the upper confidence bound on fitted-predictor
risk cannot be below the prescribed threshold
at this sample size. Specifically,
\eqref{eq:floor} gives $U_{{\rm fit},m}\geq B(1-\alpha^{1/256})
\approx0.10975>\tau$. This inequality holds even if the fitted
predictor achieves zero empirical MSE
on the calibration observations. Under the
prescribed squared-loss bound and confidence
allocation, \eqref{eq:kl-sample-size} shows
that at least 282 calibration observations
are necessary for the upper confidence bound
to meet $\tau=0.1$. A larger calibration
sample does not guarantee this outcome,
because the upper confidence bound also
depends on the observed prediction errors.

\subsection{Class Exclusion on a Fixed ECG Data Set} \label{sec:ptbxl-exclusion}

\emph{Assessment and reference.}
This experiment investigates whether any decoder
in a prescribed KAHM class can meet an empirical
MSE threshold defined using the performance of
a reference regression model. We use a fixed
PTB-XL dataset containing $N=8192$ history-target
pairs from distinct patients, comprising 4096
calibration and 4096 final-evaluation observations,
with one pair per patient
\citep{wagner2020ptbxl,wagner2022ptbxl103}.
Each history contains 32 past measurements
spanning 310 ms. The target is the lead-II
amplitude 40 ms ahead, measured in millivolts
(mV), clipped to $[-2,2]$, and divided by two.
All reported MSE values use these normalized
targets. The reference model is a histogram gradient-boosted
tree regressor trained on raw histories, with
at most 15 leaves per tree. Among candidate histogram gradient-boosted tree regressors, we select the model with the lowest
MSE on a separate validation dataset. The selected
model produces predictions clipped to $[-1,1]$. Let $r_{\rm ref}$ denote
the empirical MSE of this model on the fixed
PTB-XL dataset. We define the threshold
$\tau_{\rm PTB}$ by rounding $1.1r_{\rm ref}$
upward, giving
$\tau_{\rm PTB}\geq1.1r_{\rm ref}$ and
$\tau_{\rm PTB}\approx0.0058458874$.

\emph{Decoder class and lower bounds.}
The selected two-coordinate KAHM representation
$\Gamma$ uses coordinatewise min--max scaling,
with scaling parameters estimated from the
development data, a maximum local subspace
dimension of two, and affinity exponent
$\omega=0.5$. We consider the class
$\mathcal G_\kappa(d)$ of all measurable decoders
with predictions in $[-1,1]$ and Lipschitz
constant at most $\kappa$, where $d$ is the
maximum groupwise Euclidean distance between
feature vectors. The minimum empirical MSE
over this class is
\[
a_\kappa
=\inf_{g\in\mathcal G_\kappa(d)}
\frac1N\sum_{i=1}^N
\bigl(y_i-g(\Gamma(x_i))\bigr)^2,
\qquad \kappa\geq0.
\]
To obtain a verified lower bound on $a_\kappa$,
we partition the $N$ observations into disjoint
blocks of sizes 32, 64, and 128. For each block,
we evaluate the dual objective in
\eqref{eq:cohortdual} to obtain a lower bound
on the minimum total squared error. Summing
these lower bounds over the blocks and dividing
by $N$ gives a lower bound on $a_\kappa$
(Proposition~\ref{prop:fixed-data-risk}). We report the largest lower bound obtained
across the block sizes and the two partitioning
procedures described in
Appendix~\ref{app:protocols}. The dual objectives
are evaluated using exact rational arithmetic
and conservative upper bounds on feature
distances.

\emph{Results.}
For the selected two-coordinate KAHM representation
with $\kappa=1$, the verified lower bound on
minimum empirical MSE satisfies
$a_1\geq0.00643206>\tau_{\rm PTB}$. Consequently, no decoder in this class can meet
the empirical MSE threshold defined using the
reference regression model. For the separately selected six- and
fourteen-coordinate KAHM representations
with $\kappa=1$, the corresponding verified
lower bounds are $0.00515356$ and $0.00495880$,
respectively. Both bounds are below
$\tau_{\rm PTB}$ and therefore do not
establish class exclusion. For the two-coordinate KAHM representation, the dual calculations also establish $a_\kappa\leq\tau_{\rm PTB}
\quad\Longrightarrow\quad
\kappa\geq54.82579$. Thus, a decoder using this representation can
meet the empirical MSE threshold only if its
Lipschitz constant is at least $54.82579$
under the prescribed feature metric.

\subsection{Minimum Empirical Risk and Fitted-Predictor Error}
\label{sec:bike-class-risk}

\emph{Assessment and predictor classes.}
This experiment examines whether the empirical
MSE of fitted predictors reflects the minimum
empirical MSE achievable by their prescribed
decoder classes. Using the Bike Sharing dataset
\citep{fanaee2013bike}, we evaluate predictions
of bike rental counts for casual users,
registered users, and all users 24 hours ahead.
For each target, we evaluate $N=172$ fixed
history--target pairs using 18 feature
representations: two raw-history, four Nystr\"om,
and twelve KAHM representations. This gives
54 target-representation classes. We normalize the target values by subtracting
the minimum training target value and dividing
by the range of the training targets, separately
for each target. We define the target-specific
MSE threshold $\tau$ as $0.95$ times the
empirical MSE of a persistence predictor on
the training data. This predictor forecasts
the target value 24 hours ahead using its most
recently observed value. Because normalization
uses the minimum and range of the training
targets, normalized targets in the evaluation
dataset may lie outside $[0,1]$. These values
are retained when computing empirical MSE,
whereas decoder predictions are restricted
to $[0,1]$. For each target-representation combination $m$,
let $\mathcal G_m$ denote the class of measurable
decoders with predictions in $I=[0,1]$ and
Lipschitz constant at most $\kappa=2$ under
Euclidean distance between feature vectors.
For the stored feature vectors $z_{mi}$ and
normalized targets $y_i$, define
\[
\widehat R_m(g)
=\frac1N\sum_{i=1}^N
\bigl(y_i-g(z_{mi})\bigr)^2,
\qquad
a_m=\inf_{g\in\mathcal G_m}\widehat R_m(g).
\]
We treat the stored double-precision feature
coordinates as exact inputs to these
empirical-risk calculations.

\emph{Risk bounds and fitted predictors.}
For each decoder class $m$, we compute a verified
lower bound $\underline a_m$ on its minimum
empirical MSE using the dual formulation of
Proposition~\ref{prop:fixed-data-risk}.
We obtain an upper bound $\overline a_m$ by
constructing a prediction vector that satisfies
the prescribed output and pairwise Lipschitz
constraints. Proposition~\ref{prop:fixed-data-risk}
guarantees that such a vector corresponds to
an admissible decoder. Consequently,
$\underline a_m\leq a_m\leq\overline a_m$. Across all 54 decoder classes, the largest
difference $\overline a_m-\underline a_m$
is approximately $2.181\times10^{-14}$
in normalized MSE units. Appendix~\ref{app:protocols}
describes the numerical computations and
verification procedures. For each decoder class, we evaluate two predictors
fitted on training data: a clipped affine ridge
predictor $\widehat g_{m,1}$ and a clipped
Lipschitz-extension predictor $\widehat g_{m,2}$. For each decoder class $m$ and predictor index
$j\in\{1,2\}$, let $\underline r_{m,j}$
denote a verified numerical lower bound on the
empirical MSE $\widehat R_m(\widehat g_{m,j})$
of the corresponding fitted predictor. The difference between
the fitted predictor's empirical MSE and the
minimum empirical class risk satisfies $\widehat R_m(\widehat g_{m,j})-a_m
\geq \underline r_{m,j}-\overline a_m$. If $\underline r_{m,j}>\overline a_m$, the
empirical MSE of the fitted predictor exceeds
the minimum empirical MSE achievable by its
prescribed decoder class.
\begin{table}[tbp]
\centering\small
\begin{tabular}{@{}lrrr@{}}
\toprule
Target &
\shortstack{Classes containing\\a decoder with\\MSE below $\tau$} &
\shortstack{Classes with both\\fitted predictors'\\MSEs above $\tau$} &
\shortstack{Fitted predictors with\\MSE above their\\class minimum}\\
\midrule
Casual & 18 & 3 & 36\\
Registered & 18 & 18 & 36\\
Total rentals & 18 & 18 & 36\\
\midrule
All targets & 54 & 39 & 108\\
\bottomrule
\end{tabular}
\caption{Counts from empirical-MSE comparisons on the fixed
Bike Sharing assessment sets.
The three count columns require
$\overline a_m<\tau$,
$\min_{j=1,2}\underline r_{m,j}>\tau$, and
$\underline r_{m,j}>\overline a_m$, respectively.
Each target has 18 classes and 36 fitted predictors.
Raw total counts equal casual plus registered counts,
so the three targets are not independent replications.}
\label{tab:diagnosis}
\end{table}

\emph{Results and interpretation.}
Table~\ref{tab:diagnosis} shows that all 54
decoder classes contain an admissible decoder
whose empirical MSE is below the corresponding
threshold. However, in 39 of these classes,
the empirical MSE of each of the two fitted
predictors exceeds the threshold. Moreover,
all 108 fitted predictors have empirical MSE
at least $0.01794$ above the minimum empirical
MSE of their corresponding decoder classes,
in normalized MSE units. These findings demonstrate that the empirical
MSE achieved by fitted predictors can
overestimate the minimum empirical MSE
achievable by their decoder classes. The
feasible prediction vectors used to establish
class feasibility are constructed using the
assessment targets and therefore do not
establish predictive performance on unseen
observations.

\subsection{Forecasting Performance and Computational Cost}
\label{sec:forecasting-cost}

\emph{Screening.}
We examine whether lower bounds on minimum empirical
class risk can exclude candidate decoder classes
before fitting their predictors. On the Beijing
air-quality dataset \citep{chen2015beijing},
screening evaluates 18 candidate classes but
excludes none, so no decoder fits are avoided.
Computing the lower bounds requires 18.90 seconds
in the recorded experiment. In the three-target
Bike Sharing experiment, screening also avoids
no additional decoder fits compared with
model search without screening.

\emph{PTB-XL comparison.}
We evaluate whether the three validation-selected
KAHM predictors, using two-, six-, and
fourteen-coordinate representations, satisfy
both an empirical MSE requirement and a model
storage requirement. The predictors are evaluated
on the same 8192 PTB-XL observations used in
Section~\ref{sec:ptbxl-exclusion}. The numerical accuracy test requires an upper
bound on each predictor's empirical MSE to be
no greater than a lower bound on $1.1r_{\rm ref}$,
where $r_{\rm ref}$ is the empirical MSE of the
histogram gradient-boosted reference model.
The storage requirement is that the KAHM
predictor uses fewer stored numerical bytes
than the reference model. None of the three
KAHM predictors passes the numerical accuracy
test and, consequently, none satisfies the
joint accuracy and storage requirement.

\emph{Deployment comparison.}
We compare KAHM predictors using normalized
affinity features and clipped reconstruction-score
features. The evaluation comprises five
experimental runs using a hidden Markov model
(HMM) and one chronological forecasting
experiment using the Appliances dataset
\citep{candanedo2017appliances}. For each experiment, the conventional accuracy
reference is selected by minimizing validation
MSE. Among conventional predictors whose
validation MSE is at most $1.1$ times that
of the accuracy reference, we select the
predictor requiring the least model storage
as the cost reference. Within each KAHM
candidate family, we select the predictor
requiring the least storage subject to the
same validation-MSE criterion. Across all six evaluations, the selected
predictors from both KAHM families achieve
assessment MSE at most $1.1$ times that
of the conventional accuracy reference.
However, every selected KAHM predictor
requires more numerical model storage
and has a longer measured inference time
per observation than the corresponding
conventional cost reference.
Appendix~\ref{app:forecasting-cost} reports
the selection procedures, measurement
definitions, and numerical comparisons.

%% file: sections/discussion.tex
\section{Discussion and Limitations}
\label{sec:discussion}

\emph{Interpreting class-risk bounds.}
Equation~\eqref{eq:decomposition} expresses optimal
class risk as the sum of irreducible prediction
error given the full history, conditional-mean
information loss caused by the representation,
and approximation error imposed by the prescribed
Lipschitz decoder class. A lower bound on optimal
class risk constrains the sum of these terms but
does not identify their individual contributions. The MEM experiment illustrates this distinction.
All 16 possible histories have distinct MEM
feature vectors, so the representation retains
the information in the observed history on this
finite support. Nevertheless, the prescribed
MEM decoder class is excluded. The class
exclusion therefore reflects the Lipschitz
restriction rather than a loss of predictive
information caused by the representation.

\emph{Block size and numerical optimization.}
Although the expected block scores $\mathcal W_b$
are nondecreasing with block size, the corresponding
lower confidence bounds need not increase. For a
fixed calibration sample of size $n$, larger blocks
provide fewer independent block scores for direct
calibration through \eqref{eq:Lblock}. Moreover,
any fixed block size can yield an arbitrarily
weak lower bound relative to positive optimal
class risk (Proposition~\ref{prop:limits}).
Consequently, the theoretical hierarchy does
not determine which block size produces the
strongest finite-sample certificate
(Appendix~\ref{app:resolution}). Empirical-risk calibration offers another use
of blockwise optimization. Verified block lower
bounds can be combined into a lower bound on
minimum empirical MSE through \eqref{eq:reuse}
and calibrated using \eqref{eq:Lemp}. In the
finite-state experiment, full-sample optimization
produces the strongest lower confidence bounds.

\emph{Interpretation of fixed-dataset results.}
The PTB-XL and Bike Sharing experiments establish
deterministic conclusions about minimum empirical
MSE on the evaluated datasets. Unlike population-risk
confidence bounds, these conclusions do not require
iid calibration observations. However, they do not
establish guarantees for population MSE. In the
Bike Sharing experiment, the feasible prediction
vectors are constructed using the evaluation targets
and therefore do not demonstrate predictive
performance on unseen observations.

\emph{Sampling assumptions.}
The population-risk confidence bounds require iid
calibration observations independent of model
development, with targets and decoder predictions
in a known bounded interval. Candidate
representation-decoder classes, fitted predictors,
calibration sample sizes, confidence-bound types,
and failure-probability allocations are fixed
before calibration. Direct block calibration
also requires block sizes and partitions to be
specified in advance. Numerical optimization may nevertheless use the
calibration observations to select pairwise
Lipschitz constraints and dual multipliers.
For empirical-risk calibration, the resulting
lower bound $d_{m,n}$ must be measurable and
satisfy $0\leq d_{m,n}\leq a_{m,n}$ for every
dataset (Section~\ref{sec:calibration}).
For example, in industrial predictive maintenance,
individual machine runs may constitute iid time-series
realizations independent of the development data.
For the lower confidence bounds, sampling one prediction
origin per run as specified in Section~\ref{sec:calibration}
yields iid history-target pairs conditional on the
development data. The sample size $n$ counts independent
machine runs. History-target pairs from nonoverlapping windows
within a run need not be independent.
The stated population-risk guarantees do not directly
apply to dependent calibration observations.

\emph{Numerical validity.}
Exact evaluation of the dual objective verifies
the optimization lower bounds but does not
establish the numerical accuracy of the feature
representations used in their computation.
The finite-state demonstrator and Bike Sharing
experiments treat the stored feature coordinates
as exact inputs. The PTB-XL class-exclusion
result requires the supplied coordinate intervals
to contain the actual KAHM feature values.
In the feature-augmentation experiment, the
numerical verification does not rigorously bound
every error introduced during feature construction
and subsequent computations
(Appendix~\ref{app:feature-implementation}).

\emph{Computational cost and further work.}
Corollary~\ref{cor:decisions} provides lower and
upper bounds on the minimum assigned cost of
meeting a prescribed MSE threshold among
prespecified representation-decoder classes. These assigned costs need
not equal the actual training or deployment
costs of fitted predictors. In the reported experiments, class-risk screening
did not reduce the number of decoder fits, and the
selected KAHM predictors required more model storage
and inference time than their conventional cost
references (Section~\ref{sec:forecasting-cost}).
An important direction for further work is to
determine whether class-risk certificates can
reduce total model-selection costs after accounting
for feature construction, bound computation,
decoder fitting, and numerical verification.

\section{Conclusion}

We established a hierarchy of lower bounds on the
optimal population mean squared error of prescribed
representation-decoder classes. The expected
normalized block scores are nondecreasing and
converge to optimal class risk on compact metric
feature spaces, although any fixed block size
can yield an arbitrarily weak relative lower
bound. We combine verified numerical lower
bounds on empirical risk with statistical
calibration to obtain simultaneous lower
confidence bounds on optimal class risk and
upper confidence bounds on fitted-predictor
risk under independent iid calibration.

These bounds support two distinct decisions:
excluding every decoder in a class when its
lower confidence bound exceeds the prescribed
MSE threshold, and establishing class feasibility
when an admissible fitted predictor has an upper
confidence bound at most equal to the threshold. The
experiments demonstrate why fitted-predictor
performance alone cannot characterize the
predictive capability of a prescribed decoder
class. The fixed-dataset experiments establish
conclusions about empirical MSE, while the
reported computational comparisons do not
demonstrate savings from class-risk certification.

%% file: sections/resolution.tex
\section{Finite-Sample Accuracy of Risk Bounds}
\label{app:resolution}
We bound the approximation error introduced by
finite block sizes and quantify the effects of
calibration sample size and numerical uncertainty
on simultaneous risk bounds.
\subsection{Approximation Error from Finite Blocks}
\begin{proposition}[Covering-number bounds on the approximation error]
\label{prop:cover}
Let $(U_i,Y_i)_{i=1}^b$, $b\geq2$, be iid copies of
$(U,Y)$, with $U\in\mathcal U$ and
$Y\in I=[y_-,y_+]$ almost surely, where $I$ is bounded.
Let $d$ be a measurable, finite-valued pseudometric on
$\mathcal U$, and let $\mathcal G=\mathcal G_\kappa(d)$
be the class of all measurable $\kappa$-Lipschitz
decoders $g:\mathcal U\to I$, for fixed $\kappa\geq0$.
Set $D=y_+-y_->0$ and $B=D^2$, and define
\[
\begin{aligned}
R(g)&=\mathbb E[(Y-g(U))^2],
& R_\kappa^*&=\inf_{g\in\mathcal G}R(g), & 
a_b&=\inf_{g\in\mathcal G}\frac1b
\sum_{i=1}^b(Y_i-g(U_i))^2.
\end{aligned}
\]
Assume $a_b$ is measurable, and set $T_b=ba_b/(b-1)$,
$\mathcal W_b=\mathbb E[T_b]$, and
$\Delta_b=R_\kappa^*-\mathcal W_b$. Fix a measurable set $S\subseteq\mathcal U$ with
$\Pr(U\in S)=1$.
For $\eta>0$, let $N_{\mathcal G}(\eta)$ be the smallest
number $N\geq1$ of decoders
$g_1,\ldots,g_N\in\mathcal G$ satisfying
\[
\min_{1\leq j\leq N}\sup_{u\in S}|g(u)-g_j(u)|
\leq\eta
\qquad\text{for every }g\in\mathcal G,
\]
with $N_{\mathcal G}(\eta)=\infty$ if no finite cover
exists.
For every $\eta>0$ with $N_{\mathcal G}(\eta)<\infty$,
\begin{equation}
0\leq\Delta_b\leq
\min\left\{\frac B4,\;
2D\eta+B\sqrt{\frac{\log N_{\mathcal G}(\eta)}{2b}}\right\}.
\label{eq:cover}
\end{equation}
For $r>0$, let $M(r)$ be the smallest number of closed
radius-$r$ balls with centers in $\mathcal U$ covering
$S$, or $\infty$ if no finite cover exists.
For $\kappa>0$ and $\eta>0$, set
$J_\eta=1+\lceil2D/\eta\rceil$. Then
\begin{equation}
\log N_{\mathcal G}(\eta)
\leq M\!\left(\frac{\eta}{4\kappa}\right)\log J_\eta.
\label{eq:metric-cover}
\end{equation}
For $\kappa=0$, $\Delta_b=0$.
\end{proposition}

\begin{proof}
The proof of Theorem~\ref{thm:hierarchy} shows that the minimum sum of squared errors in \eqref{eq:block} equals $ba_b$. Write $\widehat R_b(g)=b^{-1}\sum_i(Y_i-g(U_i))^2$. Take a cover of size $N=N_{\mathcal G}(\eta)<\infty$
and set $Z_j=R(g_j)-\widehat R_b(g_j)$.
For $y,v,w\in I$,
$|(y-v)^2-(y-w)^2|\leq2D|v-w|$.
The covering property and $R(g_j)\geq R_\kappa^*$
therefore give
\[
a_b\geq\min_j\widehat R_b(g_j)-2D\eta
\geq R_\kappa^*-\max_j Z_j-2D\eta.
\]
For each fixed decoder $g_j$, the variables $R(g_j)-(Y_i-g_j(U_i))^2$, $i=1,\ldots,b$, are iid, have mean zero, and lie in an interval
of length $B$. Hoeffding's lemma therefore gives
\[
\mathbb E[e^{tZ_j}]
\leq \exp\left(\frac{t^2B^2}{8b}\right),
\qquad t>0.
\]
Using Jensen's inequality and
$e^{t\max_j Z_j}\leq\sum_{j=1}^N e^{tZ_j}$,
we obtain
\[
\begin{aligned}
\mathbb E\left[\max_{1\leq j\leq N}Z_j\right]
&\leq \inf_{t>0}\frac1t
\log\left(\sum_{j=1}^N
\mathbb E[e^{tZ_j}]\right)
&\leq \inf_{t>0}
\left\{\frac{\log N}{t}
+\frac{tB^2}{8b}\right\} 
&=B\sqrt{\frac{\log N}{2b}}.
\end{aligned}
\]
Taking expectations and using
$\mathcal W_b\geq\mathbb E[a_b]$ gives the second upper
bound in \eqref{eq:cover}.
Theorem~\ref{thm:hierarchy} and the constant midpoint
decoder give $0\leq\Delta_b\leq R_\kappa^*\leq B/4$. For \eqref{eq:metric-cover}, suppose $\kappa>0$ and
$M(r)<\infty$, where $r=\eta/(4\kappa)$.
Partition $I$ into at most $J_\eta$ intervals of width
at most $\eta/2$.
If two decoders take values in the same interval at
every cover center, the Lipschitz constraints bound
their difference on $S$ by
$2\kappa r+\eta/2=\eta$.
Selecting one decoder for each interval assignment
realized by some $g\in\mathcal G$ gives a cover with
at most $J_\eta^{M(r)}$ representatives, proving
\eqref{eq:metric-cover}.
If $M(r)=\infty$, the inequality is immediate.
For $\kappa=0$, all decoders are constant and the
empirical minimizer is the sample mean.
Thus $T_b$ is the unbiased sample variance and
$\mathcal W_b=\operatorname{Var}(Y)=R_\kappa^*$.
\end{proof}

\subsection{Finite-Sample Accuracy and Risk Comparisons}
\begin{theorem}[Accuracy of simultaneous risk bounds]
\label{thm:resolution}
For $m\in\{c,r\}$, let
$\Gamma_m:\mathcal X\to\mathcal U_m$ be measurable and
$d_m$ a measurable, finite-valued pseudometric on
$\mathcal U_m$. Let $\mathcal G_m=\mathcal G_{\kappa_m}(d_m)$ be the
class of all measurable $\kappa_m$-Lipschitz decoders
$\mathcal U_m\to I$, with $\kappa_m\geq0$ and a common
bounded interval $I=[y_-,y_+]$. The representations and decoder classes are fixed
using development data. The fitted decoder
$\widehat g_r:\mathcal U_r\to I$ is measurable and
fixed independently of the calibration data.
It need not belong to $\mathcal G_r$. Their risks use the same joint distribution of $(X,Y)$,
with $Y\in I$ almost surely and $B=(y_+-y_-)^2>0$:
\[
R_m(g)=\mathbb E[(Y-g(\Gamma_m(X)))^2],
\qquad
R_c^*=\inf_{g\in\mathcal G_c}R_c(g).
\]
The classes need not be nested. Let $n\geq b\geq2$ and $N\geq1$.
The calibration samples $(X_i,Y_i)_{i=1}^n$ and
$(X'_i,Y'_i)_{i=1}^N$ are each iid from this distribution
and independent of development data; they may overlap.
The block scores $T_k$, $k=1,\ldots,K$, use
$K=\lfloor n/b\rfloor$ disjoint blocks of $b$
observations.
Each score is the minimum total squared error under
class $c$'s pairwise Lipschitz constraints, divided by
$b-1$, as in \eqref{eq:block}.
The fitted-predictor losses are
$\ell_i=(Y'_i-\widehat g_r(\Gamma_r(X'_i)))^2$.
The sample sizes, block size, block index sets, and
probabilities $\alpha_L,\alpha_U,\beta_L,\beta_U\in(0,1)$
are fixed before examining calibration data.
Set $C_b=B\lfloor b^2/4\rfloor/[b(b-1)]$,
$\mathcal W_{c,b}=\mathbb E[T_k]$, and
$\Delta_b=R_c^*-\mathcal W_{c,b}\geq0$. Suppose measurable numerical intervals and upper bounds
$\nu_L,\nu_U$ on their mean widths satisfy, for every
calibration data set,
\[
\begin{aligned}
0&\leq\underline T_k\leq T_k\leq\overline T_k\leq C_b,
&0&\leq\underline\ell_i\leq\ell_i\leq\overline\ell_i\leq B,\\
\nu_L&\geq\frac1K\sum_{k=1}^K
(\overline T_k-\underline T_k),
&\nu_U&\geq\frac1N\sum_{i=1}^N
(\overline\ell_i-\underline\ell_i).
\end{aligned}
\]
Using $k_-$ and $k_+$ from \eqref{eq:kl}, define the
exact KL endpoints and assume the measurable reported
bounds $L,U$ satisfy
\[
\begin{aligned}
L^{\rm KL}
&=C_b k_-\!\left(
\frac{K^{-1}\sum_k\underline T_k}{C_b},
\frac{\log(1/\alpha_L)}K\right),
&L^{\rm KL}-w_L&\leq L\leq L^{\rm KL},\\
U^{\rm KL}
&=B k_+\!\left(
\frac{N^{-1}\sum_i\overline\ell_i}{B},
\frac{\log(1/\alpha_U)}N\right),
&U^{\rm KL}&\leq U\leq U^{\rm KL}+w_U.
\end{aligned}
\]
The numerical inequalities hold for every data set,
with measurable $w_L,w_U\geq0$.
Define
\[
\begin{aligned}
E_L&=C_b\left(
\sqrt{\frac{\log(1/\alpha_L)}{2K}}
+\sqrt{\frac{\log(1/\beta_L)}{2K}}
\right)+\nu_L+w_L,\\
E_U&=B\left(
\sqrt{\frac{\log(1/\alpha_U)}{2N}}
+\sqrt{\frac{\log(1/\beta_U)}{2N}}
\right)+\nu_U+w_U.
\end{aligned}
\]
All probability statements below are conditional on
development data.
With probability at least $1-\alpha_L-\alpha_U$,
$L\leq R_c^*$ and $R_r(\widehat g_r)\leq U$.
With probability at least $1-\beta_L-\beta_U$,
\begin{align}
R_c^*-L&\leq\Delta_b+E_L,
\label{eq:lower-resolution}\\
U-R_r(\widehat g_r)&\leq E_U,
\label{eq:upper-resolution}\\
L-U&\geq R_c^*-R_r(\widehat g_r)-\Delta_b-E_L-E_U.
\label{eq:gain-resolution}
\end{align}
These inequalities and the coverage inequalities hold
jointly with probability at least
$1-\alpha_L-\alpha_U-\beta_L-\beta_U$.
On this joint event,
$R_c^*-R_r(\widehat g_r)>\Delta_b+E_L+E_U$
implies $L>U$, so the fitted predictor has smaller
population risk than every decoder in $\mathcal G_c$.
For a fixed MSE threshold $\tau\geq0$, the conditions
\[
R_c^*-\tau>\Delta_b+E_L,\qquad
\tau-R_r(\widehat g_r)\geq E_U
\]
imply, on this joint event, $U\leq\tau<L$ and
$R_r(\widehat g_r)\leq\tau<R_c(g)$ for every
$g\in\mathcal G_c$.
\end{theorem}
\begin{proof}
Write
$\bar T=K^{-1}\sum_{k=1}^K T_k$ and
$\bar\ell=N^{-1}\sum_{i=1}^N\ell_i$.
Since $\underline T_k\leq T_k$ and
$\ell_i\leq\overline\ell_i$, the monotonicity
of $k_-$ and $k_+$ in their first arguments
gives, on the respective confidence events
in \eqref{eq:kl},
\[
\begin{aligned}
L
&\leq L^{\rm KL}
\leq C_b k_-\left(
\frac{\bar T}{C_b},
\frac{\log(1/\alpha_L)}{K}\right)
\leq \mathcal W_{c,b}
\leq R_c^*,\\
R_r(\widehat g_r)
&\leq B k_+\left(
\frac{\bar\ell}{B},
\frac{\log(1/\alpha_U)}{N}\right)
\leq U^{\rm KL}
\leq U.
\end{aligned}
\]
The first confidence event has probability
at least $1-\alpha_L$, and the second has
probability at least $1-\alpha_U$.
The union bound therefore establishes
simultaneous coverage with probability
at least $1-\alpha_L-\alpha_U$. The binary Pinsker inequality
$\operatorname{kl}(q\Vert p)\geq2(q-p)^2$ gives
\[
Ck_-(q,\varepsilon)\geq Cq-C\sqrt{\varepsilon/2},
\qquad
Ck_+(q,\varepsilon)\leq Cq+C\sqrt{\varepsilon/2}
\]
for $C>0$, $q\in[0,1]$, and $\varepsilon\geq0$. The interval enclosures imply
$K^{-1}\sum_k\underline T_k\geq\bar T-\nu_L$
and $N^{-1}\sum_i\overline\ell_i\leq\bar\ell+\nu_U$.
Combining these inequalities with Hoeffding's inequality
gives, jointly with probability at least
$1-\beta_L-\beta_U$,
\[
\begin{aligned}
L&\geq\bar T-\nu_L
-C_b\sqrt{\frac{\log(1/\alpha_L)}{2K}}-w_L
\geq\mathcal W_{c,b}-E_L,\\
U&\leq\bar\ell+\nu_U
+B\sqrt{\frac{\log(1/\alpha_U)}{2N}}+w_U
\leq R_r(\widehat g_r)+E_U.
\end{aligned}
\]
Substituting $\mathcal W_{c,b}=R_c^*-\Delta_b$ gives
\eqref{eq:lower-resolution} and \eqref{eq:upper-resolution};
subtracting the two endpoint bounds gives
\eqref{eq:gain-resolution}.
The union bound gives the joint probability.
The risk-gap condition makes the right-hand side of
\eqref{eq:gain-resolution} positive.
The threshold conditions give $L>\tau$ from
\eqref{eq:lower-resolution} and $U\leq\tau$ from
\eqref{eq:upper-resolution}.
On the joint event, $R_r(\widehat g_r)\leq U$ and
$R_c(g)\geq R_c^*\geq L$ for every $g\in\mathcal G_c$,
which proves the population-risk comparisons.
\end{proof}

%% file: sections/protocols.tex
\section{Experimental Protocols and Risk Bounds on Fixed Data Sets}
\label{app:protocols}

The controlled experiments use iid calibration observations
independent of development data. The PTB-XL and Bike Sharing
studies use fixed assessment data.

\subsection{Development and Calibration with iid Observations}

\emph{Demonstrator development.}
The development search in Section~\ref{sec:demonstrator} compares
KAHM models with maximum local subspace dimensions
$r_{\max}\in\{1,2\}$, keeping $\omega=16$ and
$\eta=10^{-6}$ fixed.
For RAW, MEM, and LOCAL, the Gaussian Nystr\"om readout uses
the same 16 distinct training histories as landmarks.
Representation coordinates are standardized using their
training means and standard deviations before kernel evaluation.
The resulting Nystr\"om features are standardized separately
using the training sample.
In both steps, a standard deviation below $10^{-8}$ is replaced
by one.
For standardized representation coordinates $u,v$, the kernel is
$\exp\{-\|u-v\|_2^2/(2\sigma^2)\}$.
We use $\sigma^2=\max(q^2\delta,10^{-12})$, where
$q\in\{1/4,1/2,1,2,4\}$ and $\delta$ is the median squared
distance between standardized landmark pairs with squared
distance above $10^{-14}$; $\delta=1$ if no such pair exists.
The ridge penalties are $10^{-6},10^{-5},\ldots,10^2$.
The ridge objective is mean squared training error plus the
penalty times the squared coefficient norm, with an unpenalized intercept. For each candidate maximum local subspace dimension
$r_{\max}\in\{1,2\}$, we select the Nystr\"om
bandwidth multiplier and ridge penalty separately
for the MEM and LOCAL readouts, minimizing the
validation MSE of their clipped predictions.
We then select a common value of $r_{\max}$
for both representations by minimizing the
average of their two resulting validation MSEs. RAW selects its readout independently over the same 45
bandwidth-penalty pairs.
Thus the search fits 45 RAW, 90 MEM, and 90 LOCAL predictors.
Ties prefer the smaller bandwidth multiplier, then the smaller
ridge penalty; a tie between values of $r_{\max}$ favors $r_{\max}=1$.
The selected predictors retain their training-fitted parameters.
Calibration observations and population risks under the known
law do not enter this selection.

\emph{Feature-augmentation development.}
Each of the twelve repetitions under each of the two joint
distributions of $(X,Y)$
in Section~\ref{sec:feature-augmentation-experiment} uses
disjoint iid samples: 256 observations for representation
fitting, 512 for decoder training, 256 for validation,
4096 for calibration, and 1024 for final evaluation.
The representation search uses min-max or standard scaling,
maximum local subspace dimensions $r_{\max}\in\{1,2\}$,
and affinity exponents
$\omega\in\{0.5,4,16,64\}$.
The ridge penalties are $0,10^{-4},10^{-2}$.
The six-coordinate predictor with the smallest validation
MSE determines the shared feature hyperparameters.
The two-coordinate representation retains its first feature
group, and validation selects that representation's ridge
penalty separately from the same set.
Both fitted decoders use the clipped affine family in
\eqref{eq:refinement}, with $\kappa=2$ and $I=[-0.85,0.85]$.
The selected representations and predictors remain fixed
during calibration and final evaluation.

\subsection{Minimum Empirical Risk on a Fixed Data Set}
\begin{proposition}[Deterministic bounds on minimum empirical risk]
\label{prop:fixed-data-risk}
Let $\mathcal U$ be a measurable space with a measurable
finite-valued pseudometric $d$. Fix observations
$z_i\in\mathcal U$ and $y_i\in\mathbb R$, $i=1,\ldots,M$,
with $M\geq1$.
Let $\mathcal F$ be the class of measurable $\kappa$-Lipschitz
maps from $(\mathcal U,d)$ to $I=[y_-,y_+]$, where
$\kappa\geq0$ and $-\infty<y_-\leq y_+<\infty$.
The targets may lie outside $I$.
Define $R_M(h)=M^{-1}\sum_i(h(z_i)-y_i)^2$ for
$h:\mathcal U\to\mathbb R$, and
$R_M^\star=\inf_{f\in\mathcal F}R_M(f)$.
This infimum is attained and equals
\begin{equation}
R_M^\star=\frac1M\min_{v\in I^M}
\left\{\sum_i(v_i-y_i)^2:
|v_i-v_j|\leq\kappa d(z_i,z_j)\quad(i<j)\right\}.
\label{eq:cohort}
\end{equation}
For every $i<j$, suppose
$0\leq d^-_{ij}\leq d(z_i,z_j)\leq d^+_{ij}<\infty$.
For arbitrary real multipliers $\lambda_{ij}$, $i<j$, set
$s_i=\sum_{j>i}\lambda_{ij}-\sum_{j<i}\lambda_{ji}$.
Then
\begin{equation}
D(\lambda)=
\sum_i\min_{t\in I}\{(t-y_i)^2+s_it\}
-\kappa\sum_{i<j}d^+_{ij}|\lambda_{ij}|
\leq M R_M^\star.
\label{eq:cohortdual}
\end{equation}
The $i$th scalar minimum occurs at $\clip_I(y_i-s_i/2)$. For any $v\in I^M$ satisfying
$|v_i-v_j|\leq\kappa d^-_{ij}$ for all $i<j$, define
\[
L_0=\frac1M\max\left\{
D(\lambda),\sum_i(y_i-\clip_I(y_i))^2\right\},
\qquad
U_0=\frac1M\sum_i(v_i-y_i)^2.
\]
Numerical endpoints $L\leq L_0$ and $U\geq U_0$ satisfy
$L\leq R_M^\star\leq U$.
For any predictor $h$ with $\ell_h\leq R_M(h)\leq u_h$,
\[
\ell_h-U\leq R_M(h)-R_M^\star\leq u_h-L.
\]
If $h\in\mathcal F$, then
$R_M(h)-R_M^\star\geq\max\{0,\ell_h-U\}$.
For any $\tau\geq0$, $L>\tau$ implies $R_M(f)>\tau$
for every $f\in\mathcal F$, and $U\leq\tau$ implies that
some $f\in\mathcal F$ satisfies $R_M(f)\leq\tau$.
\end{proposition}
\begin{proof}
Every $f\in\mathcal F$ gives a feasible vector in
\eqref{eq:cohort}. Conversely, a feasible vector $v$ extends to
$f_v(u)=\clip_I(\min_j\{v_j+\kappa d(u,z_j)\})$
\citep{gottlieb2017regression}.
Each function $u\mapsto v_j+\kappa d(u,z_j)$ is measurable
and $\kappa$-Lipschitz. Their finite minimum has both
properties, and clipping to $I$ is measurable and
$1$-Lipschitz. Hence $f_v\in\mathcal F$.
At $u=z_i$, every term in the minimum is at least $v_i$
by the pairwise constraints, and the $i$th term equals
$v_i$. Since $v_i\in I$, $f_v(z_i)=v_i$.
This proves \eqref{eq:cohort}.
Constant vectors in $I^M$ are feasible, and the feasible
set is compact, so the continuous objective attains
its minimum. For every $v$ feasible in \eqref{eq:cohort},
$s^\top v=\sum_{i<j}\lambda_{ij}(v_i-v_j)
\leq\kappa\sum_{i<j}d^+_{ij}|\lambda_{ij}|$.
The scalar minima defining $D(\lambda)$ are no greater
than their values at $v_i$, so
$D(\lambda)\leq\sum_i(v_i-y_i)^2$.
Minimizing over feasible $v$ proves \eqref{eq:cohortdual}.
Completing the square gives the scalar minimizers.
Setting $\lambda=0$ gives
$D(0)=\sum_i(y_i-\clip_I(y_i))^2$.
The inequalities
$|v_i-v_j|\leq\kappa d^-_{ij}\leq\kappa d(z_i,z_j)$
establish feasibility for the true distances, yielding
$L\leq L_0\leq R_M^\star\leq U_0\leq U$.
Subtracting the bounds on $R_M^\star$ from those on
$R_M(h)$ gives the predictor comparison;
$h\in\mathcal F$ additionally implies $R_M(h)\geq R_M^\star$.
The threshold implications follow from the risk bounds
and attainment.
\end{proof}
\subsection{PTB-XL and Bike Sharing Calculations}
\label{app:fixed-data-calculations}
\emph{PTB-XL.}
The assessment set combines 4096 calibration and 4096 final-evaluation
observations from distinct PTB-XL 1.0.3 patients, with one
history-target pair per patient
\citep{wagner2020ptbxl,wagner2022ptbxl103}.
For each separately selected two-, six-, or fourteen-coordinate map,
two partitions use groups of 32 observations. One follows row order;
the other repeatedly groups the lowest-index unused row with its
31 nearest unused neighbors using Euclidean distances between the
stored feature vectors, breaking ties by row index.
Within each partition, merging two or four successive groups of 32
gives groups of 64 and 128.
The class constraints use the maximum Euclidean distance over
feature groups. For each partition and block size
$b\in\{32,64,128\}$, we sum the verified
dual lower bounds on block squared errors
and divide by $N=8192$. For each representation
and prescribed Lipschitz constant, the reported
lower bound on minimum empirical MSE is the
maximum of the six resulting values. Dual objectives are evaluated exactly using distance upper bounds
from the supplied feature intervals. Validity assumes these
intervals contain the exact feature coordinates. To obtain the necessary bound on $\kappa$
reported in Section~\ref{sec:ptbxl-exclusion},
we keep the recorded dual multipliers fixed
while varying $\kappa$. For each partition
and block size, summing the dual objectives
in \eqref{eq:cohortdual} gives $Na_\kappa\geq A-\kappa P$, $P=\sum_k\sum_{i<j}d^+_{ij,k}|\lambda_{ij,k}|\geq0$, where $A$ is the sum of the terms independent of $\kappa$, and $k$ indexes the blocks. For $P>0$, this inequality implies $a_\kappa\leq\tau_{\rm PTB}
\quad\Longrightarrow\quad
\kappa\geq
\max\left\{0,\frac{A-N\tau_{\rm PTB}}{P}\right\}$. The largest such bound across the two
partitions and three block sizes for the
two-coordinate KAHM representation is
$54.82579$, after conservative downward
rounding.

\emph{Bike Sharing.}
The casual, registered, and total rental targets each have
$M=172$ observations and 18 feature maps \citep{fanaee2013bike}.
The maps combine min-max or standard scaling with 24 raw-history
coordinates, 8 or 24 Nystr\"om coordinates, or 8 or 24 KAHM
coordinates with maximum local subspace dimensions
$r_{\max}\in\{4,12,20\}$.
KAHM retains the first group of 8 coordinates or both groups of
8 and 16 coordinates.
Each class uses Euclidean distance, $\kappa=2$, and $I=[0,1]$.
Stored binary64 feature vectors $z_i$ are exact inputs;
losses use the unclipped, training-normalized targets
(Section~\ref{sec:evidence}). Distance bounds $0\leq d^-_{ij}\leq\|z_i-z_j\|_2\leq d^+_{ij}$
are multiples of $2^{-80}$, verified by exact squared comparisons.
Each initial prediction vector $q\in[0,1]^M$ is obtained by
clipping an approximate quadratic-programming solution or the
assessment targets to $[0,1]$, then rounding to multiples of
$2^{-40}$. Signed dual multipliers use the same rounding grid.
Let $\rho_{ij}$ be shortest-path distances on the complete graph
of observations with edge weights $2d^-_{ij}$. The vectors $v_i^-=\min_j\{q_j+\rho_{ij}\}$, $v_i^+=\max_j\{q_j-\rho_{ij}\}$ belong to $I^M$ and satisfy
$|v_i^\pm-v_j^\pm|\leq\rho_{ij}\leq2d^-_{ij}$. For every $t\in[0,1]$, the prediction vector
$v(t)=(1-t)v^-+tv^+$ belongs to $I^M$ and
satisfies the prescribed pairwise Lipschitz
constraints, since these constraints are
preserved under convex combinations.
Consequently, $R_M^\star
\leq
\min_{0\leq t\leq1}
\frac1M\sum_{i=1}^M
\bigl(y_i-(1-t)v_i^--tv_i^+\bigr)^2$. We evaluate this one-dimensional minimum
exactly to obtain an upper bound on the
minimum empirical MSE. The lower bound is $L_0$ in Proposition~\ref{prop:fixed-data-risk},
using $d^+_{ij}$ in the dual objective. 
The reported upper bounds on minimum empirical MSE
use initial prediction vectors $q$ obtained from
approximate quadratic-programming solutions in
52 decoder classes and from the observed targets
in the remaining two classes. To assess empirical class feasibility, we
initialize $q$ using the observed targets for
all 54 classes and apply the feasible prediction-vector
construction described above. For every class,
the resulting prediction vector satisfies the
prescribed Lipschitz constraints and achieves
empirical MSE below its target-specific threshold.
\subsection{Forecasting Comparisons and Computational Cost}
\label{app:forecasting-cost}

\emph{Screening comparisons.}
In the Beijing air-quality and Bike Sharing
experiments, we compare the number of decoder
fits required by candidate search with and
without class-risk screening. Screening excludes
a candidate decoder class before model fitting
when a verified lower bound on its minimum
empirical MSE exceeds the prescribed threshold.
Both experiments use the target normalization
and persistence-based MSE threshold defined
in Section~\ref{sec:bike-class-risk}.

\emph{PTB-XL comparison.}
The evaluation uses three KAHM predictors with
two-, six-, and fourteen-coordinate representations,
selected using validation MSE. Their empirical
MSEs are evaluated on the same 8192 PTB-XL
observations used in
Section~\ref{sec:ptbxl-exclusion}. For each selected predictor $h$, let
$\overline r(h)$ denote a numerical upper bound
on its empirical MSE, and let $\tau^-_{\rm PTB}$
denote a downward-rounded lower bound on
$1.1r_{\rm ref}$, where $r_{\rm ref}$ is the
empirical MSE of the histogram gradient-boosted
reference model. The numerical accuracy test
requires $\overline r(h)\leq\tau^-_{\rm PTB}$. To satisfy the joint accuracy and storage requirement, the predictor must additionally
use fewer stored numerical bytes than the
reference model.

\emph{Deployment selection and assessment.}
The forecast horizon is four steps for the hidden
Markov model and six steps for the Appliances dataset.
For each experiment, we select the conventional
predictor with the lowest validation MSE as the
accuracy reference $h_{\rm acc}$. Among conventional predictors whose validation
MSE is at most $1.1$ times that of $h_{\rm acc}$,
we select the predictor requiring the fewest
stored numerical bytes as the cost reference
$h_{\rm cost}$. Separately, within each KAHM
candidate family, we select the predictor
requiring the fewest stored numerical bytes
subject to the same validation-MSE criterion. For each selected KAHM predictor $h$, we
evaluate the joint accuracy, storage, and
inference-time requirements $\widehat R(h)\leq1.1\widehat R(h_{\rm acc})$, $S(h)<S(h_{\rm cost})$, $T(h)\leq T(h_{\rm cost})$, where $\widehat R$ denotes empirical MSE
computed from the stored double-precision
predictions and assessment targets.
The storage measure $S$ and inference-time
measure $T$ are defined below. Table~\ref{tab:deployment-ratios} reports
the corresponding numerical comparisons.

\emph{Storage and inference time.}
The storage measure $S$ counts retained encoder and decoder
arrays and numerical scalars in bytes.
Each timed batch uses the first 512 raw assessment histories
and includes input scaling, feature computation, and prediction.
After warm-up, nine batches are timed with single-threaded
execution. The measure $T$ is the median batch time divided
by 512.
The reproduction scripts recompute these summaries from the
stored timing measurements.
\begin{table}[tbp]
\centering\small
\begin{tabular}{@{}lrrrrrr@{}}
\toprule
& \multicolumn{3}{c}{Affinity features}
& \multicolumn{3}{c}{Clipped scores}\\
\cmidrule(lr){2-4}\cmidrule(l){5-7}
Evaluation & $r_R$ & $r_S$ & $r_T$ & $r_R$ & $r_S$ & $r_T$\\
\midrule
HMM repetition 1 & 1.081 & 258.2 & 67.8 & 1.080 & 258.2 & 56.6\\
HMM repetition 2 & 1.074 & 326.1 & 81.5 & 1.062 & 326.1 & 68.3\\
HMM repetition 3 & 1.083 & 258.2 & 60.9 & 1.081 & 258.2 & 48.8\\
HMM repetition 4 & 1.062 & 258.2 & 58.9 & 1.062 & 258.2 & 52.2\\
HMM repetition 5 & 1.085 & 258.2 & 65.9 & 1.081 & 258.2 & 51.9\\
Appliances & 1.063 & 535.5 & 163.6 & 0.994 & 535.5 & 160.5\\
\bottomrule
\end{tabular}
\caption{Deployment comparisons for normalized affinity
features and clipped reconstruction-score features.
The ratios are
$r_R=\widehat R(h)/\widehat R(h_{\rm acc})$,
$r_S=S(h)/S(h_{\rm cost})$, and
$r_T=T(h)/T(h_{\rm cost})$.
The required conditions are $r_R\leq1.1$, $r_S<1$,
and $r_T\leq1$.
Ratios are rounded for display; comparisons use the
unrounded recorded values.}
\label{tab:deployment-ratios}
\end{table}

\subsection{Feature Computation and Numerical Conventions}
\label{app:feature-implementation}
The feature-augmentation experiment in
Section~\ref{sec:feature-augmentation-experiment}
uses the KAHM implementation documented in the
\href{https://github.com/MohitKumarRostock/Kernel_Affine_Hull_Koopman_Machine/blob/8baad4931f8a934c4b09612e7e8f90149873a1ca/docs/FEATURE_COMPUTATION.md}{\underline{standalone feature-computation reference}}. Local subspaces are constructed by principal component
analysis (PCA). The parameter $r_{\max}$ specifies the maximum local subspace dimension, $d_x$ the input dimension,
and $N_{\rm loc}$ the number of reference vectors after
subsampling or augmentation. The initial requested subspace
dimension is $\min(r_{\max},d_x,N_{\rm loc}-1)$.
Only leading covariance eigenvectors with computed eigenvalues
above $2^{-52}$ are retained. The requested subspace
dimension is reduced until all projected reference coordinates
have range at least $10^{-3}$ or the requested dimension
reaches one. A fit with no retained direction fails. Local reconstructions are computed as
kernel-weighted averages of the reference vectors.
The interval evaluation of each reconstruction
requires an interval for the sum of kernel weights
that excludes zero. If this condition is not
established, the arithmetic precision is increased.
Unresolved evaluations are reported as failures
without discarding observations. For the numerical feature values, we evaluate
the normalized affinity formula with
$\eta=10^{-6}$ at the midpoints of the
reconstruction-score intervals using 160-bit
arithmetic. The resulting feature coordinates
are stored in double precision. Separately,
we propagate the full reconstruction-score
intervals through the normalized affinity
formula to obtain intervals containing the
exact feature coordinates for the stored
inputs and model parameters. The lower
interval endpoints are rounded downward
and the upper endpoints upward.